# A QoE-Aware Split Inference Accelerating Algorithm for NOMA-based Edge Intelligence

Xin Yuan, Ning Li, *Member*, *IEEE*, Quan Chen, Wenchao Xu, Zhaoxin Zhang, Song Guo, *Fellow, IEEE*

*Abstract*—Even the AI has been widely used and significantly changed our life, deploying the large AI models on resource limited edge devices directly is not appropriate. Thus, the model split inference is proposed to improve the performance of edge intelligence (EI), in which the AI model is divided into different sub-models and the resource-intensive sub-model is offloaded to edge server wirelessly for reducing resource requirements and inference latency. However, the previous works mainly concentrate on improving and optimizing the system's QoS, ignore the effect of QoE which is another critical item for the users except for QoS. Even the QoE has been widely learned in EC, considering the differences between task offloading in EC and split inference in EI, and the specific issues in QoE which are still not addressed in EC and EI, these algorithms cannot work effectively in edge split inference scenarios. Thus, an effective resource allocation algorithm is proposed in this paper, for 1) accelerating split inference in EI and 2) achieving the tradeoff between inference delay, QoE, and resource consumption, abbreviated as ERA. Specifically, the ERA takes the resource consumption, QoE, and inference latency into account to find the optimal model split strategy and resource allocation strategy. Since the minimum inference delay and resource consumption, and maximum QoE cannot be satisfied simultaneously, the gradient descent (GD) based algorithm is adopted to find the optimal tradeoff between them. Moreover, the loop iteration GD approach (Li-GD) is developed to reduce the complexity of the GD algorithm caused by parameter discretization. Additionally, the properties of the proposed algorithms are investigated, including convergence, complexity, and approximation error. The experimental results demonstrate that the performance of ERA is much better than that of the previous studies.

*Index Terms*—Edge Intelligence; Model Split; Inference Accelerating; QoE

## I. INTRODUCTION

Artificial intelligence (AI) has been widely used and has significantly changed our lifestyle, such as metaverse [1-2], automatic driving [2-4], image generation [5], etc. However, because the sizes of AI models are always large to provide high-performance services, the quantity of computing resources required for these models is also large. Therefore, it is inappropriate to deploy these AI models on edge devices, such as mobile phones, vehicles, unmanned aerial vehicles (UAVs), etc., in which the computing resources are quite limited. To address this issue, one possible solution is to split the large AI model into different sub-models and offload the resource-intensive part to the edge server wirelessly (e.g., WiFi, 5G, etc.) to reduce resource requirement and inference latency [6-11]. The model split strategies between device, edge server and cloud has been investigated by previous works, and some excellent works have been proposed, such as [12-19]. These studies find the optimal model segmentation points or early-exist points to minimize inference delay and resource requirements while maintaining high inference accuracy by reinforcement learning [12], convex optimization [13-16], heuristic algorithm [17-19], etc.

However, these algorithms mainly concentrate on improving and optimizing the system's Quality of Service (QoS), such as low inference latency and energy consumption, high inference accuracy, etc., ignore the effect of Quality of Experience (QoE) which is one of the critical items for users except for QoS. Even the QoE has been widely learned in edge computing, such as [20-33], considering the differences[1] between task offloading in edge computing and split inference in edge intelligence [12-19, 34], these algorithms cannot work effectively in edge split inference scenarios. Additionally, due to the following concerns and issues in QoE, which are still not fully addressed in edge computing, the performance of edge split inference can be improved further.

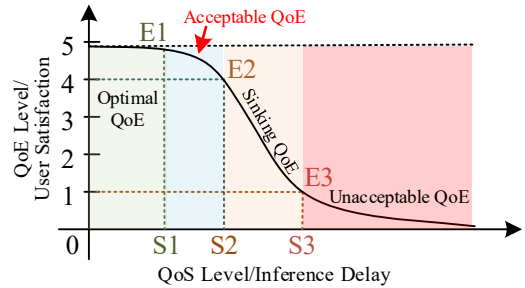

Fig.1. The relationship of QoS and QoE

First, as shown in Fig.1, for individual user, the users' QoE and the inference delay follow a sigmoid-like curve [21][23]. Therefore, the relationship between QoE and inference delay can be categorized into four sensitivity classes roughly [21][23]. *Optimal QoE* (0~S1): QoE is not sensitive to the inference delay so that a slight growth of the delay would not affect the QoE distinctively. *Acceptable QoE* (S1~S2): QoE reduces with the increasing of inference delay but still can satisfy most users' requirements. *Sinking QoE* (S2~S3): QoE is sensitive to the service delay. Specifically, as the delay grows, the QoE sinks rapidly. *Unacceptable QoE* (S3~∞): The QoE is too low to be accepted by the users. The Fig.1 means that the users' QoE does not immediately drop to zero even if the delay slightly exceeds the predefined threshold, i.e., S1, and instead decreases gradually as the delay increases. According to this observation, it is reasonable to relax the predefined threshold of QoE from Optimal QoE (S1) to Acceptable QoE (S2) to reduce the huge amount of resource consumption which is used to maintain low inference delay, and guarantee the performance of users' QoE.

---

[1] The model splitting in edge intelligence is more fine-grained than the task offloading in edge computing, which will increase the complexity and difficulty of model splitting compared with the task offloading in edge computing.

Second, traditionally, the existing works achieve high-performance edge split inference through minimize the sum of all the user's inference delay[2] [12-18, 35]. Since they do not consider the users' QoE, the QoE of the whole system cannot guarantee. Besides, even low inference delay means high QoE for individual user, this is not applicable for the whole system. For instance, as shown in Fig.2, the green bar represents the inference latency threshold of Optimal QoE, i.e., S1. When the inference delay is larger than the green bar, the QoE will deteriorate; when the inference delay is smaller than the green bar, the users' QoE increment is too slight to be noticed by the users, which is not necessary and wastes network resource. The blue bar represents the inference delay when considering the users' QoE, and the red bar is the inference delay without considering the users' QoE. As a result, on the one hand, the sum of blue bar is 9 + 18 + 4 + 15 = 46 and the sum of red bar is 11 + 5 + 7 + 20 = 43, which means that the inference delay of the whole system when considering the QoE is larger than that without considering the QoE. On the other hand, all the blue bars are smaller than the green bar which means that all the users' QoE are satisfied under this strategy. However, the red bars of user1, user3, and user4 are all larger than the green bar, and the sum of the exceeded value over S1 is 9. Under this strategy, only 25% users' QoE requirements is satisfied. This phenomenon indicates that due to the heterogeneous QoE requirements of users, low inference delay cannot guarantee high-performance QoE of the whole system.

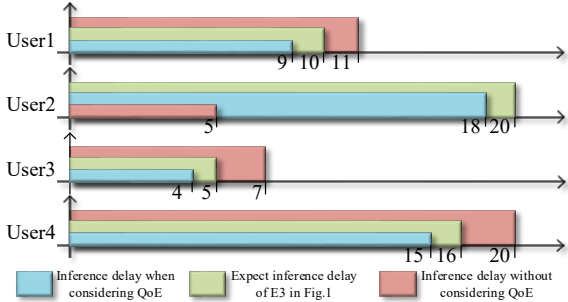

Fig.2. An example for the relationship of inference delay and QoE

These two observations indicate that it is possible and reasonable to reduce a large amount resource consumption while maintain high-performance QoE for edge split inference by relaxing the requirements on inference latency. The reasons can be explained as follows. First, maintaining low inference delay needs to consume large amount of computing and communication resource. Thus, reducing the requirements on inference delay can save resource significantly. This has a great significance in EI, in which the resource is quite limited. Second, for the individual users, slightly reducing the requirements on inference delay may not cause serious QoE reduction; for the whole system, considering the users' QoE can improve the performance of the whole system. Therefore, it provides a potential solution for the existing dilemma in EI to achieve low resource consumption and high-performance QoE simultaneously.

Based on the above analysis, for accelerating split inference in EI and achieving the tradeoff between inference delay, QoE, and resource consumption, A QoE-Aware Split Inference Accelerating Algorithm for NOMA-based Edge Intelligence is proposed in this paper, abbreviated as ERA. Specifically, when the user has a model inference task that needs to be calculated in the edge server, it takes resource consumption, the inference delay, and the QoE into account to find the optimal model split strategy, subchannel allocation strategy in NOMA, transmission power allocation strategy, and computing resource allocation strategy. Since the minimum inference delay, maximum QoE, and minimum resource consumption cannot be satisfied simultaneously, the gradient descent (GD) algorithm is adopted to determine the optimal tradeoff between them. Moreover, considering that the variables of subchannel allocation, model split strategy, the parameters of QoE are discrete, the loop iteration GD approach (Li-GD) is proposed to reduce the complexity of the GD algorithm caused by discrete parameters. The key idea of the Li-GD algorithm is that: the initial value of the $i$th layer's GD procedure is selected from the optimal results of the former $(i-1)$ layers' GD procedure whose intermediate data size is the closest to the $i$th layer. Additionally, the properties of the proposed algorithms are investigated, including convergence, complexity, and approximation error.

The main contributions of this paper can be summarized as follows.

1) To the best of our knowledge, this is the first work which considers the user's QoE in edge split inference. In this paper, the two observations presented in Fig.1 and Fig.2 make it is possible and reasonable to reduce a large amount resource consumption while maintain high-performance QoE for edge split inference by relaxing the requirements on inference latency. Based on this conclusion, we propose the joint optimization problem between minimum inference delay, maximum QoE, and minimum resource consumption for edge split inference. The purpose is to find the optimal resource allocation strategy, transmission power strategy and model split strategy for the above joint optimization problem to achieve high-performance edge split inference in EI.

2) In this paper, because minimum inference delay, maximum QoE, and minimum resource consumption cannot be achieved simultaneously, the GD-based algorithm is adopted in this study to effectively achieve an optimal tradeoff between them. Moreover, considering the complexity of this issue caused by uneven and discrete intermediate data size, we propose a Li-GD algorithm to improve the efficiency of the GD procedure. The key idea of the Li-GD algorithm is that: the initial value of the ith layer's GD procedure is selected from the optimal results of the former $(i-1)$ layers' GD procedure whose intermediate data size is the closest to the ith layer.

---

[2] In previous works, since minimizing the inference delay of each user is NP-hardness, the utility function is always to minimize the sum of all the users' inference delay [12-18, 35].

3) The properties of the proposed Li-GD algorithm are investigated. The Li-GD algorithm is convergent, and the convergence time is $K = \frac{\|x^0 - x^*\|_2^2}{2\eta\epsilon}$, the complexity of the Li-GD is $O(X\bar{K}\mathcal{F}Mx^3 \ln^2(x))$, the approximate error is smaller than $\frac{\varepsilon}{\rho_{min}(1-B_{max})\log_2\left(1 + \frac{P_{min}}{\Delta^* + \frac{\alpha P_{max}}{2}}\right)}$. Additionally, it can reduce the complexity and convergence time compared with the traditional GD approach.

The remainder of this paper is organized as follows. The network models and the problems to be solved in this paper are presented in Section II. The Li-GD algorithm is proposed in Section III, and its properties, e.g., the convergence, the complexity, etc., are also investigated in this section. In Section IV, the effectiveness of the proposed ERA algorithm is demonstrated through simulation. Finally, Section V summarizes the conclusions of this work.

## II. NETWORK MODEL AND PROBLEM STATEMENT

In this section, the split inference, the inference delay, the QoE model and the resource consumption models (including the computing and communication resource) that used in this paper are introduced. The same as our previous work [34], the network model is illustrated in Fig.3. Subsequently, based on the proposed models, the problems that will be solved in this study are also described in detail in this section.

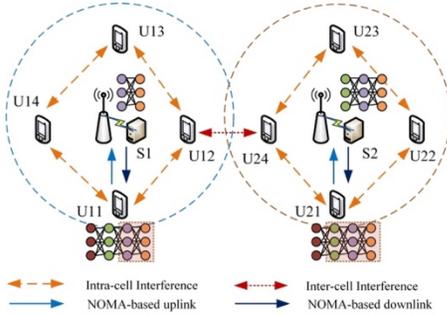

Fig.3. Network model.

As shown in Fig.3, we consider mobile edge computing in a multi-cell network with N single-antenna APs and U single-antenna end devices, indexed by $N = \{1,2,...,N\}$ and $U = \{1,2,...,U\}$, respectively. The NOMA technology is applied for both uplink and downlink during the data transmission in this paper. Thus, the total system bandwidth B is equally divided into M orthogonal subchannels, indexed by $M = \{1,2,...,M\}$. We consider the use of the nearest AP association policy [48]. As such, each end device associates with the nearest AP that can provide the maximum average channel gain. We denote the set of devices served by AP n as $U_n$. Therefore, each device $u \in U$ offloads its split inference model to its associated AP $n \in N$ via subchannel $m \in M$. We assume that $M < N$ and denote the set of devices served by AP n on subchannel m as $U_n^m$. Therefore, the devices associated with different APs may access to the same subchannel and interfere with each other. The uplink and downlink between mobile user and edge server are all NOMA-based channels. Thus, not only can the users in the coverage of the same AP interfere with each other, but also with the users who select the same subchannel in adjacent APs. For each mobile user, there is a large model with $\mathcal{F}$ layers needs to be split and offloaded from device to edge server.

### A. Split inference model

In this section, we take YOLOv2 as an example to demonstrate the model split policy and the effect of different model split strategies on communication overhead. As shown in Fig.4 [34], the model has 16 available split points, which are $s_1, s_2, ..., s_{15}, s_{16}$, respectively. Note that $s_1$ means the entire model is offloaded to and calculated on the edge server, $s_{16}$ means that the entire model is calculated on the device. The number of layers of YOLOv2 is larger than that of available split points. This means that not all the layers in the model can be selected as split points. Therefore, for model $\mathcal{M}$, its model split policies can be expressed as: $S^{\mathcal{M}} = s_i^{\mathcal{M}} (\forall i \in \{1,2,...,\mathcal{F}_{\mathcal{M}}\})$, where $\mathcal{F}_{\mathcal{M}}$ is the number of available split points in model $\mathcal{M}$, in which $s_1$ means the whole model is offloaded to and calculated on edge server, $s_{\mathcal{F}_{\mathcal{M}}}$ means the whole model is calculated on device. For instance, in Fig.1, $\mathcal{F}_{\mathcal{M}} = 16$ and $s_3$ means the model split point is between layer Max1 and Convn2. Therefore, when the model split policy is $s_i$, it means that the first to $s_i$-th layers are calculated on mobile device, and the $(s_i + 1)$-th to $\mathcal{F}_{\mathcal{M}}$-th layers are offloaded to the edge server for deep inference. The intermediate data between $s_i$-th layer and $(s_i + 1)$-th is transmitted through wireless channel.

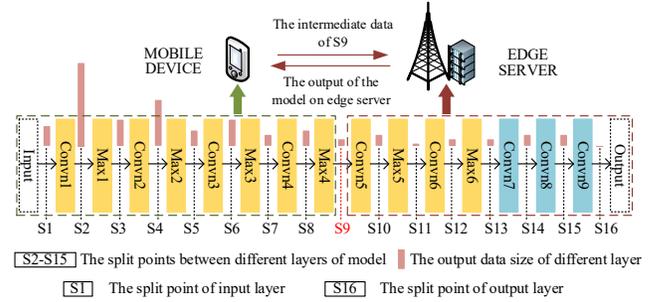

Fig.4. Split inference model.

Therefore, the intermediate data size can affect the transmission delay and inference delay seriously. For instance, as shown in Fig.4, the intermediate data size between Convn1 and Max1 is about 50 times larger than that between Max5 and Convn6. Thus, if the split point is selected as $s_1$ rather than $s_{10}$, the data transmission delay will be serious. However, this does not mean that the smaller intermediate data size, the lower inference delay is. Because the inference delay is also affected by the calculation time of the sub-models on edge server and devices, which is decided by the computing capability of edge server and devices, respectively. The details of the inference delay model and transmission delay model are presented below.

### B. Inference delay model

The inference delay includes three parts: 1) the delay that caused by model inference on mobile device, 2) the delay that caused by model inference on edge server, and 3) the delay that caused by intermediate data transmission between the device and edge server.

**(1) Inference delay on device**

As presented in Section II.A, the number of model layers is $\mathcal{F}$ and the model split decision is $s_i$, which means that the first to $s_i$-th layers are calculated on mobile device i, and the $(s_i +$ 1)-th to $\mathcal{F}$-th layers are offloaded to the edge server for deep inference. $c_i$ is defined as the floating-point operation capability of device i. Then, the inference latency on mobile devices after model segmentation can be calculated as:

$$T_i^{device} = \sum_{\delta=1}^{s_i} \frac{f_{l_\delta}}{c_i} \quad (1)$$

where $f_{l_\delta}$ is the computation task of each layer in the main branch, containing convolutional layer $f_{conv}$, pooling layer $f_{pool}$, and ReLU layer $f_{relu}$ [12]. Thus, $f_{l_\delta}$ can be computed as:

$$f_{l_\delta} = m_{\delta 1} f_{conv} + m_{\delta 2} f_{pool} + m_{\delta 3} f_{relu} \quad (2)$$

where $m_{\delta 1}$, $m_{\delta 2}$, and $m_{\delta 3}$ denote the number of convolutional layers, pooling layers, and ReLU layers, respectively, and $m_{\delta 1} + m_{\delta 2} + m_{\delta 3} = s_i$.

**(2) Inference delay on edge server**

The execution time is not proportional to the amount of allocated computational resources for the inference tasks, such as DNN, under the scenario that the edge server is multicore CPU. As demonstrated in [18], up to 44% error in execution time between theory and experiment. Thus, in this paper, let $r_i$ represent the number of minimum computational resource units that allocated to user i; $c_{min}$ implies the capability of minimum computational resource unit. Since in a multicore CPU scenario, the execution time is not linear with respect to the amount of allocated computational resource, a compensation function $\lambda(r_i)$ is introduced to fit the execution time in the multicore CPU scenario. For the single core scenario, the $\lambda(r_i)$ is degenerated to $r_i$, and for the multicore scenario, $\lambda(r_i) > r_i$. The $\lambda(r_i)$ can be estimated based on the approach that proposed in [18]. Therefore, in this paper, we only assume that $\lambda(r_i)$ increases with $r_i$, but not linearly. To model the nonlinearity in the execution time, the execution time on edge can be expressed as:

$$T_i^{server} = \sum_{\delta=s_i+1}^{\mathcal{F}} \frac{f_{e_\delta}}{\lambda(r_i) c_{min}} \quad (3)$$

where $f_{e_\delta}$ is the computation task of each layer in the main branch, containing convolutional layer $f_{conv}$, pooling layer $f_{pool}$, and ReLU layer $f_{relu}$; $\lambda(r_i) c_{min}$ is the amount of computing resource in edge server j that allocated to user i. Thus, $f_{e_\delta}$ can be calculated as:

$$f_{e_\delta} = m_{\delta 4} f_{conv} + m_{\delta 5} f_{pool} + m_{\delta 6} f_{relu} \quad (4)$$

where $m_{\delta 4}$, $m_{\delta 5}$, and $m_{\delta 6}$ denote the number of convolutional layers, pooling layers, and ReLU layers, respectively, and $m_{\delta 4} + m_{\delta 5} + m_{\delta 6} = \mathcal{F} - s_i$.

**(3) Network transmission delay**

There are two different types of network transmission delay: 1) the intermediate output transmission delay from the device to the edge server in the uplink and 2) the final result transmission delay from the edge server to the device in the downlink.

Firstly, for the intermediate data transmission delay, when the model is split at $s_i$-th layer, the intermediate data generated by the $s_i$-th layer will be transmitted to edge server to complete the inference. As shown in Fig.1, the end devices served by the same AP n on the same subchannel m form a NOMA cluster $U_n^m$, where NOMA protocol is adopted to perform the model offloading. Let $N_n^m$ and $h_{n,i}^m$ denote the number of devices in $U_n^m$ and the channel coefficient of the uplink from device i to AP n on subchannel m, respectively. Without loss of generality, we assume that $|h_{n,1}^m|^2 > |h_{n,2}^m|^2 > \cdots > |h_{n,i}^m|^2 > \cdots > |h_{n,N}^m|^2$. According to the rules of NOMA protocol, the APs apply SIC for multi-user detection. Specifically, each AP n sequentially decodes the signal from devices with higher channel gains and regards all the other signals as the interference. As such, the received signal-to-interference-plus-noise ratio (SINR) of AP n for device i $\in U_n^m$ is given by:

$$\Upsilon_{n,i}^m = \frac{p_{n,i}^m |h_{n,i}^m|^2}{\sum_{v=i+1}^{U} \beta_{n,v}^m p_{n,v}^m |h_{n,v}^m|^2 + \sum_{l=1,l\neq n}^{N} \sum_{t=1}^{U} \beta_{l,t}^m p_{l,t}^m |g_{l,t}^m|^2 + \sigma^2} \quad (5)$$

where $p_{n,i}^m$ is the transmission power of device i $\in U_n^m$, $g_{l,t}^m$ is the channel coefficient of the interference link from device t served by AP l except AP n on the same subchannel m, $\sigma^2$ is the additive white Gaussian noise. Additionally, for successfully channel decoding with SIC through subchannel m of AP n, let $I_n^m$ be the threshold of signal strength, only the users that satisfy the constraint $p_{n,i}^m |h_{n,i}^m|^2 > I_n^m$, $\forall i \in U_n^m$ can offload the partial model to edge server; otherwise, the entire model will be calculated on devices. Moreover, $\sum_{v=i+1}^{U} \beta_{n,v}^m p_{n,v}^m |h_{n,v}^m|^2$ is the intra-cell interference and $\sum_{l=1,l\neq n}^{N} \sum_{t=1}^{U} \beta_{l,t}^m p_{l,t}^m |g_{l,t}^m|^2$ is the inter-cell interference. Then, based on (5), the transmission rate of device i $\in U_n^m$ can be expressed as:

$$R_{n,i}^m = \beta_{n,i}^m \cdot \frac{B_{up}}{M} \log_2(1 + \Upsilon_{n,i}^m) \quad (6)$$

where $\beta_{n,i}^m$ represents the subchannel allocation variable. Specifically, $\beta_{n,i}^m = 1$ indicates that subchannel m is allocated to device i; otherwise, $\beta_{n,i}^m = 0$. Let $w_{s_i}$ represent the data size at the $s_i$-th layer. Then, the intermediate output transmission delay can be calculated as:

$$T_i^{tran-i} = \frac{w_{s_i}}{R_{n,i}^m} = \frac{w_{s_i}}{\beta_{n,i}^m \frac{B_{up}}{M} \log_2(1 + \Upsilon_{n,i}^m)} \quad (7)$$

For the final data transmission from edge server to user with downlink NOMA scheme, the AP transmits a superposition-coded signal to each user on a subchannel [36]. In downlink transmissions, users employ SIC to decode the received superposed signal. Without loss of generality, suppose that the devices served by the same AP j on the same subchannel k are represented by $U_j^k$. Let $N_j^k$ and $H_{j,i}^k$ denote the number of devices in $U_j^k$ and the channel coefficient of the downlink from edge server j to device i. Similar to the uplink, we assume that $|H_{j,1}^k|^2 < |H_{j,2}^k|^2 < \cdots < |H_{j,i}^k|^2 < \cdots < |H_{j,N}^k|^2$. For user 1, since it has the weakest channel coefficient, it decodes the superposed signal from edge server j without performing SIC. Then the user 1′s decoded component is subtracted from the superposed signal. The subsequent user in $U_j^k$, i.e., user 2, can decode the received signal without interference from user 1. Following this principle, the signal received by user i $\in U_j^k$ on subchannel k in BS j has a SINR $\Psi_{j,i}^k$ of:

$$\Psi_{j,i}^k = \frac{|H_{j,i}^k|^2 P_{j,i}^k}{\sum_{q=i+1}^{U} \beta_{j,q}^k P_{j,q}^k |H_{j,q}^k|^2 + \sum_{x=1,x\neq j}^{N} \sum_{y=1}^{U} \beta_{x,y}^k P_{x,y}^k |G_{x,y}^k|^2 + \sigma^2} \quad (8)$$

where $|H_{j,i}^k|^2$ is channel gain of user i on subchannel k, $\sum_{q=i+1}^{U} \beta_{j,q}^k P_{j,q}^k |H_{j,q}^k|^2$ is the intra-cell interference experienced by user i, $\sum_{x=1,x\neq j}^{N} \sum_{y=1}^{U} \beta_{x,y}^k P_{x,y}^k |G_{x,y}^k|^2$ is the inter-cell interference experienced by user i (caused by the neighbor AP of users i), and $\sigma^2$ is the addictive white Gaussian noise. Additionally, in (8), let $|H_{j,i}^k|^2 P_{j,i}^k > I_j^k$, $\forall i \in U_j^k$, where $I_j^k$ is the threshold of signal

strength for successful channel decoding with SIC by subchannel k of AP j. In the downlink NOMA, considering the factors that affect the condition of subchannel, the SIC decoding order of users on subchannel j must be the weaker users (high inter-cell interference and low channel gain) decode before stronger users. According to [37], the achievable data rate of user i by SIC can be expressed as:

$$\Phi_{j,i}^k = \beta_{j,i}^k \frac{B_{down}}{M} \log_2(1 + \Psi_{j,i}^k) \quad (9)$$

Let $m_i$ denotes the data size of the final inference result at edge server, then the final result transmission delay can be expressed by:

$$T_i^{tran-f} = \frac{m_i}{\Phi_i^k} = \frac{m_i}{\beta_{j,i}^k \frac{B_{down}}{M} \log_2(1+\Psi_{j,i}^k)} \quad (10)$$

Thus, the network transmission delay can be expressed as:

$$T_i^{trans} = T_i^{tran-i} + T_i^{tran-f} = \frac{w_{s_i}}{R_{n,i}^m} + \frac{m_i}{\Phi_{j,i}^k} \quad (11)$$

Intuitively, the overall execution latency of the task in mobile user i can be expressed as:

$$T_i = T_i^{device} + T_i^{server} + T_i^{trans}$$
$$= \sum_{\delta=1}^{s_i} \frac{f_{l_\delta}}{c_i} + \sum_{\delta=s_i+1}^{\mathcal{F}} \frac{f_{e_\delta}}{\lambda(r_i)c_{min}} + \frac{w_{s_i}}{R_{n,i}^m} + \frac{m_i}{\Phi_{j,i}^k} \quad (12)$$

### C. Parameters of QoE

Different users have different QoE requirements. Even for the same user, their QoE requirements on different AI applications are also different. In this paper, we use two parameters to evaluate the performance of user's QoE: 1) the sum of delayed completion time (DCT) $\mathcal{C}$ and 2) the number of users whose value of DCT is larger than zero $z$. The delayed completion time is defined in Definition 1. Moreover, in this section, we use $Q_i^m$ to represent the QoE threshold of model $m$ in user i. The QoE threshold is the inference delay (S2) of Acceptable QoE in Fig.1.

**Definition 1**. The Delayed Completion Time is the time that the actual task finish time over the expected finish time which is S2 in Fig.1., i.e., the inference delay of Acceptable QoE.

Based on Definition 1 and (12), the DCT of model $m$ in user i, i.e., $\mathcal{C}_i^m$ can be expressed as:

$$\mathcal{C}_i^m = \begin{cases} 0, & \text{if } T_i^m < Q_i^m \\ T_i^m - Q_i^m, & \text{if } T_i^m > Q_i^m \end{cases} \quad (13)$$

where $Q_i^\mathcal{M}$ is the QoE threshold of model $\mathcal{M}$ in user i, i.e., the expected finish time; the $T_i^\mathcal{M}$ is the actual finish time which can be calculated based on (12).

As shown in (13), when $T_i^\mathcal{M} < Q_i^\mathcal{M}$, $\mathcal{C}_i^m = 0$, when $T_i^\mathcal{M} > Q_i^\mathcal{M}$, $\mathcal{C}_i^m = T_i^\mathcal{M} - Q_i^\mathcal{M}$. Therefore, on the one hand, the value of $\mathcal{C}_i^m$ is discrete; on the other hand, the value of $\mathcal{C}_i^m$ is highly related to the model split strategy, the transmission power, and resource allocation strategy. Thus, it is difficult to be predicted. For solving these issues, we then transfer the expression of $\mathcal{C}_i^m$ into:

$$\mathcal{C}_i^{m\prime} = (T_i^\mathcal{M} - Q_i^\mathcal{M}) \cdot \mathcal{R}_i(T_i^\mathcal{M}) \quad (14)$$

In (14), let:

$$\mathcal{R}_i(x) = \frac{1}{1+e^{-a(x-1)}} \quad (15)$$

where $x = T_i^\mathcal{M}/Q_i^\mathcal{M}$ and $x > 0$. The $\mathcal{R}_i(x)$ is shown in Fig.5, and the values of $\mathcal{R}_i$ under different values of a are also presented in Fig.5. As shown in Fig.5, with the increasing of a, the value of $\mathcal{R}_i \in (0\ 1)$ is close to the two-value function $\psi \in [0,1]$ and the error between $\mathcal{R}_i$ and $\psi$ reduces. Additionally, this error is small enough to be ignored. For instance, when a = 2000, when $Q_i^\mathcal{M} = 10ms$, if $T_i^\mathcal{M}$ is 10.02ms, then $x = T_i^\mathcal{M}/Q_i^\mathcal{M} = 1.002$ and $\mathcal{R}_i(x) = 0.9827$. Thus, the error between $\mathcal{R}_i(x)$ and 1 is only 0.0173. This means that the value of $\mathcal{R}_i$ is close to 1 enough.

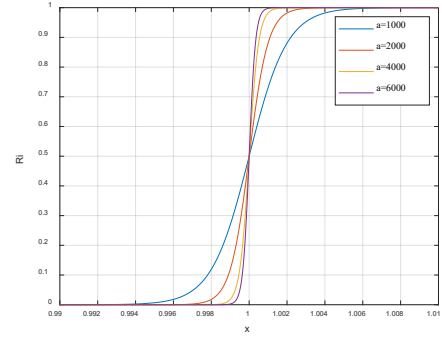

Fig.5. The value of $\mathcal{R}_i$ under different values of a.

Therefore, based on the above analysis, since $\mathcal{C}_i^m$ is discrete, according to the property of $\mathcal{R}_i$ (which is shown in Fig.5), we give the approximate rules for $\mathcal{R}_i$ as: if $\mathcal{R}_i < \frac{1}{2}$, $\mathcal{R}_i \leftarrow 0$; otherwise, if $\mathcal{R}_i > \frac{1}{2}$, $\mathcal{R}_i \leftarrow 1$. Based on this approximation, we can transfer the discrete $\mathcal{C}_i^m$ to continuous $\mathcal{C}_i^{m\prime}$ with sufficient small error. The approximation rate is decided by the value of a. The larger value of a, the small approximation error is.

Therefore, let $x = T_i^\mathcal{M}/Q_i^\mathcal{M}$, the sum of DCT of the whole system can be calculated as:

$$\mathcal{C} = \sum_{i=1}^U \mathcal{C}_i^{m\prime} = \sum_{i=1}^U (T_i^\mathcal{M} - Q_i^\mathcal{M}) \cdot \mathcal{R}_i(x)$$
$$= \frac{(T_i^\mathcal{M} - Q_i^\mathcal{M})}{1+e^{-a(T_i^\mathcal{M}/Q_i^\mathcal{M}-1)}} \quad (16)$$

Except for the $\mathcal{C}$, we also consider the number of users whose value of DCT is larger than 0, denoted as $z$. Based on (16), let $x = T_i^\mathcal{M}/Q_i^\mathcal{M}$, $z$ can be calculated as:

$$z = \sum_{i=1}^U \mathcal{R}_i(x) = \sum_{i=1}^U \frac{1}{1+e^{-a(T_i^\mathcal{M}/Q_i^\mathcal{M}-1)}} \quad (17)$$

### D. Energy consumption model

The energy consumption that considered in this paper includes: 1) the energy that consumed for model inference at end device and edge server, and 2) the energy that consumed for intermediate data transmission between end device and edge server.

**(1) Energy consumption on device**

Let $\xi_i$ represent the effective switched capacitance of CPU, which is determined by the chip structure of mobile device i, then the energy consumption of mobile devices that caused for model inference can be calculated as:

$$E_i^l = \sum_{\delta=1}^{s_i} \xi_i c_i^2 \varphi_i f_{l_\delta} \quad (18)$$

where $\varphi_i$ is the required CPU cycles to compute 1-bit data on end device.

Let $p_{n,i}^m$ denotes the transmission power of mobile device i on subchannel m, the energy consumption of intermediate data transmission can be computed as:

$$E_i^t = p_{n,i}^m \cdot \frac{w_{s_i}}{R_{n,i}^m} \quad (19)$$

**(2) Energy consumption on edge server**

For edge server, its energy consumption also comes from two aspects: the model inference and final result transmission. Based on (10), the energy consumption of intermediate data transmission can be computed as:

$$E_e^t = P_{j,i}^k \cdot \frac{m_i}{\Phi_{j,i}^k} \quad (20)$$

Let $\xi_e$ represent the effective switched capacitance of CPU, which is determined by the chip structure of edge server $e \in N$, then the energy consumption of edge server that used for model inference can be calculated as:

$$E_e^l = \sum_{\delta=s_i+1}^{\mathcal{F}} \xi_e (\lambda(r_i) c_{min})^2 \varphi_e f_{l_\delta} \quad (21)$$

where $\varphi_e$ is the required CPU cycles to compute 1-bit data on edge server.

Thus, the energy consumption for task execution and intermediate data transmission between mobile device and edge server can be derived by:

$$\begin{aligned} E_i &= E_i^l + E_i^t + E_e^l + E_e^t \\ &= \sum_{\delta=1}^{s_i} \xi_i c_i^2 \varphi_i f_{l_\delta} + \sum_{\delta=s_i+1}^{\mathcal{F}} \xi_e (\lambda(r_i) c_{min})^2 \varphi_e f_{l_\delta} \\ &\quad + p_{n,i}^m \cdot \frac{w_{s_i}}{R_{n,i}^m} + P_{j,i}^k \cdot \frac{m_i}{\Phi_{j,i}^k} \end{aligned} \quad (22)$$

*E. Problem statement*

As demonstrated in Section II.A, the split strategy of model $\mathcal{M}$ is $S^\mathcal{M} = s_i^\mathcal{M}(\forall i \in \{1,2,\ldots,\mathcal{F}_\mathcal{M}\})$; the intermediate data size that relates to different split strategies is $D^\mathcal{M} = \{d_1^\mathcal{M}, d_2^\mathcal{M}, \ldots, d_{\mathcal{F}_\mathcal{M}}^\mathcal{M}\}$. Therefore, the split inference delay includes four parts: the inference delay on device $T_i^{device}(s_i)$, the inference delay on edge server $T_i^{server}(s_i, r_i)$, the delay for intermediate data transmission $T_i^{tran-i}(s_i, \beta_{n,i}^m, p_{n,i}^m)$, and the delay for final result transmission $T_i^{tran-f}(\beta_{j,i}^k, P_{j,i}^k)$. Similarly, the energy consumption in model split inference also includes four parts: the energy consumption on device $E_i^l(s_i)$, the energy consumption for intermediate data transmission $E_i^t(s_i, \beta_{n,i}^m, p_{n,i}^m)$, the energy consumption for final result transmission $E_e^t(\beta_{j,i}^k, P_{j,i}^k)$, and the energy consumption on edge server $E_e^l(s_i, r_i)$. Moreover, the sum of DCT of the network is $\mathcal{C}(s_i, \beta_{n,i}^m, \beta_{j,i}^k, p_{n,i}^m, P_{j,i}^k, r_i)$ and the number of users whose value of DCT is larger than 0 is $z(s_i, \beta_{n,i}^m, \beta_{j,i}^k, p_{n,i}^m, P_{j,i}^k, r_i)$.

Then, based on (1), (3), (7) and (10), we obtain the inference delay: $T_i(s_i, \beta_{n,i}^m, \beta_{j,i}^k, p_{n,i}^m, P_{j,i}^k, r_i) = T_i^{device}(s_i) + T_i^{server}(s_i, r_i) + T_i^{tran-i}(s_i, \beta_{n,i}^m, p_{n,i}^m) + T_i^{tran-f}(\beta_{j,i}^k, P_{j,i}^k)$; based on (18), (19), (20) and (21), we get the energy consumption: $E_i(s_i, \beta_{n,i}^m, \beta_{j,i}^k, p_{n,i}^m, P_{j,i}^k, r_i) = E_i^l(s_i) + E_i^t(s_i, \beta_{n,i}^m, p_{n,i}^m) + E_e^t(\beta_{j,i}^k, P_{j,i}^k) + E_e^l(s_i, r_i)$. Base on (16) and (17), we obtain the sum of DCT of the whole network and the number of users whose DCT is larger than 0, i.e., $\mathcal{C}$ and $z$, respectively.

Our purpose is to achieve minimum inference delay, resource consumption, DCT, the number of users whose DCT is larger than 0, and energy consumption at the same time, with the variables are the model split strategy $S^\mathcal{M} = s_i^\mathcal{M}(\forall i \in \{1,2,\ldots,\mathcal{F}_\mathcal{M}\})$, the subchannel allocation strategy of user $i$ $B_i = \{\beta_{n,i}^m, \beta_{j,i}^k\}$, the computing resource allocation strategy $r_i \in [r_{min}\ r_{max}]$, and the transmission power allocation strategy of device $i$ $P_i = \{p_{n,i}^m, P_{j,i}^k\}$. During these variables, $S^\mathcal{M}$ and $B_i$ are discrete, while $P_i$ and $r_i$ are continuous. Therefore, the problem (P0) that to be solved in this paper can be expressed as:

$$\min\{\sum_{i=1}^U T_i, \mathcal{C}, z, \sum_{i=1}^U E_i, \sum_{i=1}^U \lambda_i\} \quad (23)$$
$$\text{s. t.} \quad S^\mathcal{M} = s_i^\mathcal{M}(\forall i \in \{1,2,\ldots,\mathcal{F}_\mathcal{M}\}) \quad (23.a)$$
$$D^\mathcal{M} = \{d_1^\mathcal{M}, d_2^\mathcal{M}, \ldots, d_{\mathcal{F}_\mathcal{M}}^\mathcal{M}\} \quad (23.b)$$
$$B_i \in \{0,1\} \quad (23.c)$$
$$p_{min} \leq P_i \leq p_{max}, \forall i \in [1\ U] \quad (23.d)$$
$$r_{min} \leq r_i \leq r_{max}, \forall i \in [1\ U] \quad (23.e)$$
$$\sum_{n=1}^N \sum_{m=1}^M \beta_{n,i}^m = 1, \forall i \in [1\ U] \quad (23.f)$$
$$\sum_{j=1}^N \sum_{k=1}^M \beta_{j,i}^k = 1, \forall i \in [1\ U] \quad (23.g)$$

In P0, the constraints of (23.a), (23.b), (23.c), (23.d), and (23.e) are easy to understand; the constraints (23.f) and (23.g) mean that each user can select only one edge server and one sub-channel in one time slot. The P0 is difficult to address, since these two optimal objectives are opposite, as demonstrated in Section I. This indicates that simultaneously finding the minimum energy consumption and minimum inference delay is impossible. Thus, in this paper, we need to find an approach to achieve an optimal tradeoff between these two objectives.

III. OPTIMAL MODEL SPLIT AND RESOURCE ALLOCATION ALGORITHM

In this section, we discuss the optimal solutions for P0. Since P0 is difficult to solve, we propose an Li-GD algorithm for P0. Moreover, the properties of the proposed Li-GD algorithm are also investigated in this section.

*A. Loop iteration GD algorithm*

Since the optimization objectives shown in P0 are opposite, we introduce the weight-based approach to construct the utility function for each mobile user that contains both these objectives, which can be expressed as:

$$U_i = \omega_T T_i + \omega_Q(\mathcal{C} + z) + \omega_R(E_i + \lambda_i) \quad (24)$$

where $\omega_T$, $\omega_R$, and $\omega_Q$ are the weights of inference delay, resource consumption, and QoE, respectively, with the constraint is $\omega_T + \omega_R + \omega_Q = 1$. The weight represents the importance of each optimal objective to users.

Additionally, the $\omega_T$, $\omega_R$, and $\omega_Q$ are hyper-parameters, which can be decided by mobile users according to their dynamic QoS requirements. For instance, if the inference delay is more important to mobile user than resource consumption, then the mobile user can set $\omega_T > \omega_R$. This approach is flexible and practicable due to the following reasons. First, in practice, the QoS requirements of the same mobile user may change depending on the dynamic environment. For instance, when the mobile devices have sufficient resource, the inference delay may be the primary factor to be considered; otherwise, when the application requires low inference latency, regardless of whether the resource consumption is high or not, the weight of inference delay should be large. Second, the QoS and QoE requirements of various mobile users and applications may be different. For instance, the resource in user A is richer than that in user B, then the weigh of $\omega_R$ in user A may be smaller than that in user B; if one application has strict restriction on QoE, then the weight of $\omega_Q$ in this user could be larger than the other users. Thus, this approach can be adjusted according to the dynamic QoS and QoE requirements flexibly.

According to (24), the P0 can be expressed as:

$$\min \sum_{i=1}^U U_i \quad (25)$$

Let $\Gamma = \sum_{i=1}^U U_i$, based on (8), (12) and (21), $\Gamma$ can be described as:

$$\Gamma = \sum_{i=1}^U \omega_T T_i(s_i, B_i, P_i, r_i) + \sum_{i=1}^U \omega_R^i(E_i(s_i, B_i, P_i, r_i) + \lambda(r_i))$$

$$+ \sum_{i=1}^{U} \omega_Q \bigl( \mathcal{C}_i(s_i, B_i, P_i, r_i) + \mathcal{R}_i(s_i, B_i, P_i, r_i) \bigr)$$

$$= \sum_{i=1}^{U} \omega_T^i \left( \sum_{\delta=1}^{s_i} \frac{f_{l_\delta}}{c_i} + \sum_{\delta=s_i+1}^{\mathcal{F}} \frac{f_{e_\delta}}{\lambda(r_i)c_{min}} + \frac{w_{s_i}}{\beta_{n,i}^m \frac{B_{up}}{M} \log_2(1+Y_{n,i}^m)} + \frac{m_i}{\beta_{j,i}^k \frac{B_{down}}{M} \log_2(1+\Psi_{j,i}^k)} \right) + \sum_{i=1}^{U} \omega_R^i \Biggl( \sum_{\delta=1}^{s_i} \xi_i c_i^2 \varphi_i f_{l_\delta} +$$

$$\sum_{\delta=s_i+1}^{\mathcal{F}} \xi_e (\lambda(r_i)c_{min})^2 \varphi_e f_{l_\delta} + p_{n,i}^m \cdot \frac{w_{s_i}}{\beta_{n,i}^m \frac{B_{up}}{M} \log_2(1+Y_{n,i}^m)} + P_{j,i}^k \cdot$$

$$\frac{m_i}{\beta_{j,i}^k \frac{B_{down}}{M} \log_2(1+\Psi_{j,i}^k)} + \lambda(r_i) \Biggr) + \sum_{i=1}^{U} \omega_Q \cdot \frac{\bigl(T_i(s_i,B_i,P_i,r_i) - Q_i^{\mathcal{M}}\bigr)}{1 + e^{-a\bigl(T_i(s_i,B_i,P_i,r_i)/Q_i^{\mathcal{M}} - 1\bigr)}} +$$

$$\sum_{i=1}^{U} \omega_Q \cdot \frac{1}{1 + e^{-a\bigl(T_i(s_i,B_i,P_i,r_i)/Q_i^{\mathcal{M}} - 1\bigr)}} \tag{26}$$

where $R_{n,i}^m$ and $\Phi_{j,i}^k$ are presented in (6) and (9), respectively.

In (26), the size of the tasks that calculated on the mobile device and edge server relates to $s_i$, i.e., $\sum_{\delta=1}^{s_i} f_{l_\delta}$ and $\sum_{\delta=s_i+1}^{\mathcal{F}} f_{e_\delta}$, and $w_{s_i}$ relates to the intermediate data that transmitted between the end device and edge server, which cannot be relaxed as continuous variables. Therefore, we define two variables as follows. For user i, let $f_l^{i-\delta} = \sum_{\delta=1}^{s_i} f_{l_\delta}$ and $s_i \in \{1,2,\dots,\mathcal{F}\}$, then $f_l^{i-1} = f_{l_1}$, $f_l^{i-2} = f_{l_1} + f_{l_2}$, and so on. Then we can change the variable $s_i$ to $f_l^i$, and $f_l^i \in \{f_l^{i-1}, f_l^{i-2}, \dots, f_l^{i-\mathcal{F}}\}$, where $f_{l_\delta}$ is calculated based on (2). Let $Z_i = \sum_{\delta=1}^{\mathcal{F}} f_{l_\delta}$ is the size of all layers, then $f_e^i = Z_i - f_l^i$. Moreover, $f_l^i$, $f_e^i$, and $w_{s_i}$ are calculated by mobile users in advance and stored in devices with inference model.

We then introduce $f_l^i$, $f_e^i$, and $w_{s_i}$ into (23). For different layers of the model, we can obtain a series of utility functions $\mathbf{\Gamma} = \{\Gamma_1, \Gamma_2, \dots, \Gamma_{s_i}, \dots, \Gamma_{\mathcal{F}}\}$, where $\Gamma_{s_i}$ can be expressed as:

$$\Gamma_{s_i} = \sum_{i=1}^{U} \omega_T^i \left( \frac{f_l^i}{c_i} + \frac{f_e^i}{\lambda(r_i)c_{min}} + \frac{w_{s_i}}{R_{n,i}^m} + \frac{m_i}{\Phi_{j,i}^k} \right)$$

$$+ \sum_{i=1}^{U} \omega_R^i \left( \xi_i c_i^2 \varphi_i f_l^i + \xi_e(\lambda(r_i)c_{min})^2 \varphi_e f_e^i + p_{n,i}^m \cdot \frac{w_{s_i}}{R_{n,i}^m} + P_{j,i}^k \cdot \frac{m_i}{\Phi_{j,i}^k} + \lambda(r_i) \right) + \sum_{i=1}^{U} \omega_Q \left( \frac{\Bigl(\bigl(\frac{f_l^i}{c_i} + \frac{f_e^i}{\lambda(r_i)c_{min}} + \frac{w_{s_i}}{R_{n,i}^m} + \frac{m_i}{\Phi_{j,i}^k}\bigr) - Q_i^{\mathcal{M}}\Bigr)}{1 + e^{-a\bigl(\bigl(\frac{f_l^i}{c_i} + \frac{f_e^i}{\lambda(r_i)c_{min}} + \frac{w_{s_i}}{R_{n,i}^m} + \frac{m_i}{\Phi_{j,i}^k}\bigr)/Q_i^{\mathcal{M}} - 1\bigr)}} \right)$$

$$+ \sum_{i=1}^{U} \omega_Q \frac{1}{1 + e^{-a\bigl(\bigl(\frac{f_l^i}{c_i} + \frac{f_e^i}{\lambda(r_i)c_{min}} + \frac{w_{s_i}}{R_{n,i}^m} + \frac{m_i}{\Phi_{j,i}^k}\bigr)/Q_i^{\mathcal{M}} - 1\bigr)}} \tag{27}$$

where $f_l^i$, $f_e^i$, and $w_{s_i}$ are already known in advance for each inference model in mobile device.

In (27), there are U mobile users. For each mobile user, we need to calculate the optimal B, P, and r. Since the parameters P and r are continuous, the variable spaces of P and r are large and infinite dimensional. Additionally, since B, P, and r are all related to $s_i$, and B and P are close coupled, it is difficult to calculate the optimal value of B, P, and r separately. Thus, in this study, for finding the optimal tradeoff between inference delay, resource consumption, and QoE, we introduce the gradient descent approach into our algorithm. To use the gradient descent to address above issue, we first need to prove that the weight function shown in (26) is differentiable.

**Definition 2 [41].** For $f(x,y,z)$, if its partial derivative on x, y, and z, i.e., $f'(x,y,z)|_x$, $f'(x,y,z)|_y$, and $f'(x,y,z)|_z$, exist and continue, the $f(x,y,z)$ is differentiable.

Based on Definition 1, we have the conclusion as follows.

**Corollary 1.** When the values of $f_l^i$, $f_e^i$, and $w_{s_i}$ are know in advance, and we loose the constraints of $\beta_{n,i}^m \in \{0,1\}$ and $\beta_{j,i}^k \in \{0,1\}$ to $\beta_{n,i}^m \in [0\ 1]$ and $\beta_{j,i}^k \in [0\ 1]$, the utility function shown in (26) is differentiable.

Proof. Since the values of $f_l^i$, $f_e^i$, and $w_{s_i}$ are already know in advance for every inference model in each mobile device, the values of $y_1 = \frac{f_l^i}{c_i}$ and $y_2 = \xi_i c_i^2 \varphi_i f_l^i$ can be calculated easily and nothing to do with the partial derivative on $B_i$, $P_i$, and $r_i$. Therefore, the partial derivative of $\Gamma_{s_i}$ on $B_i$, $P_i$, and $r_i$ can be expressed as:

$$\Gamma'_{s_i}|_{\beta_{n,i}^m} = \sum_{i=1}^{U} \left(\omega_T^i \cdot \frac{w_{s_i}}{R_{n,i}^m}\right)\bigg|_{\beta_{n,i}^m} + \sum_{i=1}^{U} \left(\omega_E^i \cdot p_{n,i}^m \cdot \frac{w_{s_i}}{R_{n,i}^m}\right)\bigg|_{\beta_{n,i}^m} \tag{28}$$

$$\Gamma'_{s_i}|_{\beta_{j,i}^k} = \sum_{i=1}^{U} \left(\omega_T^i \cdot \frac{m_i}{\Phi_{j,i}^k}\right)\bigg|_{\beta_{j,i}^k} + \sum_{i=1}^{U} \left(\omega_E^i \cdot P_{j,i}^k \cdot \frac{m_i}{\Phi_{j,i}^k}\right)\bigg|_{\beta_{j,i}^k} \tag{29}$$

$$\Gamma'_{s_i}|_{p_{n,i}^m} = \sum_{i=1}^{U} \left(\omega_T^i \cdot \frac{w_{s_i}}{R_{n,i}^m}\right)\bigg|_{p_{n,i}^m} + \sum_{i=1}^{U} \left(\omega_E^i \cdot p_{n,i}^m \cdot \frac{w_{s_i}}{R_{n,i}^m}\right)\bigg|_{p_{n,i}^m} \tag{30}$$

$$\Gamma'_{s_i}|_{P_{j,i}^k} = \sum_{i=1}^{U} \left(\omega_T^i \cdot \frac{m_i}{\Phi_{j,i}^k}\right)\bigg|_{P_{j,i}^k} + \sum_{i=1}^{U} \left(\omega_E^i \cdot P_{j,i}^k \cdot \frac{m_i}{\Phi_{j,i}^k}\right)\bigg|_{P_{j,i}^k} \tag{31}$$

$$\Gamma'_{s_i}|_{r_i} = \sum_{i=1}^{U} \left(\omega_T^i \cdot \frac{f_e^i}{\lambda(r_i)c_{min}}\right)\bigg|_{r_i} + \sum_{i=1}^{U} \bigl(\omega_E^i \cdot \xi_e(\lambda(r_i)c_{min})^2 \varphi_e f_e^i\bigr)\bigg|_{r_i} \tag{32}$$

Moreover, according to (28), let $\Delta = \sum_{l=1,l\neq n}^{N} \sum_{t=1}^{U} \beta_{l,t}^m p_{l,t}^m |g_{l,t}^m|^2 + \sigma^2$, we have:

$$\sum_{i=1}^{U} \left(\omega_T^i \cdot \frac{w_{s_i}}{R_{n,i}^m}\right)\bigg|_{\beta_{n,i}^m} = \omega_T^1 \frac{w_{s_i}}{R_{n,1}^m}\bigg|_{\beta_{n,i}^m} + \omega_T^2 \frac{w_{s_i}}{R_{n,2}^m}\bigg|_{\beta_{n,i}^m} + \cdots + \omega_T^i \frac{w_{s_i}}{R_{n,i}^m}\bigg|_{\beta_{n,i}^m} + \cdots + \omega_T^U \frac{w_{s_i}}{R_{n,U}^m}\bigg|_{\beta_{n,i}^m} \tag{33}$$

$$\omega_T^i \frac{w_{s_i}}{R_{n,i}^m}\bigg|_{\beta_{n,i}^m} = -\omega_T^i w_{s_i} \frac{\frac{B_{up}}{M}\log_2\left(1 + \frac{p_{n,i}^m|h_{n,i}^m|^2}{\sum_{v=1,v\neq i}^{N} \beta_{n,v}^m p_{n,v}^m|h_{n,v}^m|^2 + \Delta}\right)}{\left\{\beta_{n,i}^m \frac{B_{up}}{M}\log_2\left(1 + \frac{p_{n,i}^m|h_{n,i}^m|^2}{\sum_{v=1,v\neq i}^{N} \beta_{n,v}^m p_{n,v}^m|h_{n,v}^m|^2 + \Delta}\right)\right\}^2}$$

$$= -\omega_T^1 w_{s_i} \frac{1}{(\beta_{n,i}^m)^2 \left\{\frac{B_{up}}{M}\log_2\left(1 + \frac{p_{n,i}^m|h_{n,i}^m|^2}{\sum_{v=1,v\neq i}^{N} \beta_{n,v}^m p_{n,v}^m|h_{n,v}^m|^2 + \Delta}\right)\right\}^2} \tag{34}$$

$$\sum_{\tau=1,\tau\neq i}^{U} \left(\omega_T^\tau \cdot \frac{w_{s_i}}{R_{n,\tau}^m}\right)\bigg|_{\beta_{n,\tau}^m} = \omega_T^1 \frac{w_{s_i}}{R_{n,1}^m}\bigg|_{\beta_{n,1}^m} + \cdots + \omega_T^{i-1} \frac{w_{s_i}}{R_{n,i-1}^m}\bigg|_{\beta_{n,i-1}^m} + \omega_T^{i+1} \frac{w_{s_i}}{R_{n,i+1}^m}\bigg|_{\beta_{n,i+1}^m} + \cdots + \omega_T^U \frac{w_{s_i}}{R_{n,U}^m}\bigg|_{\beta_{n,U}^m}$$

$$= \sum_{\tau=1,\tau\neq i}^{U} \frac{\beta_{n,\tau}^m \frac{B_{up}}{M}}{\left(1 + \frac{p_{n,\tau}^m|h_{n,\tau}^m|^2}{\sum_{v=1,v\neq\tau}^{N} \beta_{n,v}^m p_{n,v}^m|h_{n,v}^m|^2 + \Delta}\right)\ln 2} \cdot \frac{p_{n,\tau}^m|h_{n,\tau}^m|^2 \cdot p_{n,i}^m|h_{n,i}^m|^2}{\left(\sum_{v=1,v\neq\tau}^{N} \beta_{n,v}^m p_{n,v}^m|h_{n,v}^m|^2 + \Delta\right)^2} \tag{35}$$

The calculation of $\sum_{i=1}^{U} \left(\omega_E^i \cdot p_{n,i}^m \cdot \frac{w_{s_i}}{R_{n,i}^m}\right)\bigg|_{\beta_{n,i}^m}$, $\Gamma'_{s_i}|_{\beta_{j,i}^k}$, $\Gamma'_{s_i}|_{p_{n,i}^m}$, and $\Gamma'_{s_i}|_{P_{j,i}^k}$ are similar with that shown in (34). For $\Gamma'_{s_i}|_{\beta_{n,i}^m}$ shown in (28) and $\forall \beta_{n,i}^m \in [0\ 1]$, we have $x_1 = -\frac{\omega_T^1 w_{s_i}}{(\beta_{n,i}^m)^2 \left\{\frac{B_{up}}{M}\log_2\left(1 + \frac{p_{n,i}^m|h_{n,i}^m|^2}{\sum_{v=1,v\neq i}^{N} \beta_{n,v}^m p_{n,v}^m|h_{n,v}^m|^2 + \Delta}\right)\right\}^2}$ is continuous, $x_2 = \sum_{\tau=1,\tau\neq i}^{U} \frac{\beta_{n,\tau}^m \frac{B_{up}}{M}}{\left(1 + \frac{p_{n,\tau}^m|h_{n,\tau}^m|^2}{\sum_{v=1,v\neq\tau}^{N} \beta_{n,v}^m p_{n,v}^m|h_{n,v}^m|^2 + \Delta}\right)\ln 2} \cdot \frac{p_{n,\tau}^m|h_{n,\tau}^m|^2 \cdot p_{n,i}^m|h_{n,i}^m|^2}{\left(\sum_{v=1,v\neq\tau}^{N} \beta_{n,v}^m p_{n,v}^m|h_{n,v}^m|^2 + \Delta\right)^2}$ is continuous, and $x_3 = \sum_{i=1}^{U} \left(\omega_E^i \cdot p_{n,i}^m \cdot \frac{w_{s_i}}{R_{n,i}^m}\right)\bigg|_{\beta_{n,i}^m}$ is continuous.

Thus, based on the operational rule of continuous function, the $\Gamma'_{s_i}|_{\beta^m_{n,i}}$ is continuous with $\forall \beta^m_{n,i} \in [0\ 1]$. Similarly, the $\Gamma'_{s_i}|_{\beta^k_{j,i}}$, $\Gamma'_{s_i}|_{p^m_{n,i}}$, and $\Gamma'_{s_i}|_{P^k_{j,i}}$ are all continuous. Additionally, since $\lambda(r_i)$ is continuous, $\Gamma'_{s_i}|_{r_i}$ is continuous with $\forall r_i \in [r_{min}\ r_{max}]$.

Moreover, for $\mathcal{Q}^*_i = (T^{\mathcal{M}}_i - Q^{\mathcal{M}}_i)\frac{1}{1+e^{-a\left(\frac{T^{\mathcal{M}}_i}{Q^{\mathcal{M}}_i}-1\right)}}$, let $\mathbb{Q} = \sum_{i=1}^{U}\mathcal{Q}^*_i$, then according to (33) and let $f(x) = T^{\mathcal{M}}_i$ and $g(x) = \frac{1}{1+e^{-a(x-1)}}$, then $g'(f(x)) = f'(x) \cdot g'(x)$. Since $T'_{s_i}|_{\beta^m_{n,i}}$ is continuous with $\forall \beta^m_{n,i} \in [0\ 1]$ (i.e.,$f'(x)|_{\beta^m_{n,i}}$ and the sigmoid function $g'(x) = \frac{ax}{1+e^{-a(x-1)}}$ is also continuous (as shown in Fig.5), based on the operational rule of continuous function $\mathbb{Q}'_{s_i}|_{\beta^m_{n,i}}$ is also continuous with $\forall \beta^m_{n,i} \in [0\ 1]$. Similarly, the $\mathbb{Q}'_{s_i}|_{\beta^k_{j,i}}$, $\mathbb{Q}'_{s_i}|_{p^m_{n,i}}$, and $\mathbb{Q}'_{s_i}|_{P^k_{j,i}}$ are all continuous. Thus Corollary 1 is proved. ∎

The Corollary 1 means that when the values of $f^i_l$, $f^i_e$, and $w_{s_i}$ are known, the GD approach can be used in (27) to find the optimal strategies of $B$, $P$, and $r$. However, the utility function shown in (27) is only the utility when the model segmentation point is $s_i$, there are $\mathcal{F}$ layers in the inference model, which means that the GD algorithm needs to be repeated $\mathcal{F}$ times to find the global optimal solutions for $B$, $P$, and $r$. However, considering the complexity and convergence time of GD approach, repeating the GD approach $\mathcal{F}$ times will cause serious delay and complexity. Fortunately, for the GD approach, if we can select the initial value carefully, the complexity and convergence time can be significantly reduced. Therefore, in this study, based on the greedy approach, we propose the Loop iteration GD algorithm, which is referred to as Li-GD. The details of the Li-GD algorithm are presented in TABLE I.

TABLE I. The proposed Li-GD algorithm

| **Algorithm 1:** Loop iteration GD algorithm (Li-GD) |
|---|
| **Input:** |
| Objective function: $\mathbf{\Gamma} = \{\Gamma_1, \Gamma_2, ..., \Gamma_{s_i}, ..., \Gamma_{\mathcal{F}}\}$; |
| Gradient function: $\mathbf{\nabla} = \{\nabla_{B_i} = \frac{\partial \Gamma_{s_i}}{\partial B_i}, \nabla_{P_i} = \frac{\partial \Gamma_{s_i}}{\partial P_i}, \nabla_{r_i} = \frac{\partial \Gamma_{s_i}}{\partial r_i}\}$; |
| Algorithm accuracy: ε; |
| Step size: λ; |
| **Output:** |
| The optimal solution $\mathbf{O}^* = \{\mathbf{B}^*, \mathbf{P}^*, \mathbf{r}^*\}$; |
| 1. Let $\mathbf{B}^{j(0)} \in [0\ 1]$, $P^{j(0)} \in [P_{min}\ P_{max}]$, and $\mathbf{r}^{j(0)} \in [r_{min}\ r_{max}]$, $\forall i \in [1\ U]$ and $\forall j \in [1\ \mathcal{F}]$; |
| *# Calculating the optimal strategy for the first layer #* |
| 2. If $j = 1$; |
| 3. Let $k \leftarrow 0$, $\mathbf{B}^{j(k)} = \{B^{j(k)}_1, ..., B^{j(k)}_{\mathcal{F}}\}$, $\mathbf{P}^{j(k)} = \{P^{j(k)}_1, ..., P^{j(k)}_{\mathcal{F}}\}$ and $\mathbf{r}^{j(k)} = \{r^{j(k)}_1, ..., r^{j(k)}_{\mathcal{F}}\}$; |
| 4. Calculating $\Gamma_{s_i}(\mathbf{B}^{j(k)}, \mathbf{P}^{j(k)}, \mathbf{r}^{j(k)})$; |
| 5. Calculating the gradient $\mathbf{g}_k = g(\mathbf{B}^{j(k)}, \mathbf{P}^{j(k)}, \mathbf{r}^{j(k)})$; |
| 6. If $\|\mathbf{g}_k\| < \varepsilon$, then $\mathbf{B}^{j*} \leftarrow \mathbf{B}^{j(k)}$, $\mathbf{P}^{j*} \leftarrow \mathbf{P}^{j(k)}$ and $\mathbf{r}^{j*} \leftarrow \mathbf{r}^{j(k)}$; |
| 7. Otherwise, let $\boldsymbol{\zeta}_k = -g(\mathbf{B}^{j(k)}, \mathbf{P}^{j(k)}, \mathbf{r}^{j(k)})$, and let $\mathbf{B}^{j(k+1)} = \mathbf{B}^{j(k)} + \lambda\boldsymbol{\zeta}_k$, $\mathbf{P}^{j(k+1)} = \mathbf{P}^{j(k)} + \lambda\boldsymbol{\zeta}_k$, and $\mathbf{r}^{j(k+1)} = \mathbf{r}^{j(k)} + \lambda\boldsymbol{\zeta}_k$; |
| 8. Calculating $\Gamma_{s_i}(\mathbf{B}^{j(k+1)}, \mathbf{P}^{j(k+1)}, \mathbf{r}^{j(k+1)}) = \Gamma_{s_i}(\mathbf{B}^{j(k)} + \lambda\mathbf{p}_k, \mathbf{P}^{j(k)} + \lambda\boldsymbol{\zeta}_k, \mathbf{r}^{j(k)} + \lambda\mathbf{p}_k)$; |
| 9. If $\|\Gamma_{s_i}(\mathbf{B}^{j(k+1)}, \mathbf{P}^{j(k+1)}, \mathbf{r}^{j(k+1)}) - \Gamma_{s_i}(\mathbf{B}^{j(k)}, \mathbf{P}^{j(k)}, \mathbf{r}^{j(k)})\| < \varepsilon$ or $\max\{\|\mathbf{B}^{j(k+1)} - \mathbf{B}^{j(k)}\|, \|\mathbf{P}^{j(k+1)} - \mathbf{P}^{j(k)}\|, \|\mathbf{r}^{j(k+1)} - \mathbf{r}^{j(k)}\|\} < \varepsilon$; |
| 10. then $\mathbf{B}^{j*} \leftarrow \mathbf{B}^{j(k+1)}$, $\mathbf{P}^{j*} \leftarrow \mathbf{P}^{j(k+1)}$, and $\mathbf{r}^{j*} \leftarrow \mathbf{r}^{j(k+1)}$; |
| 11. otherwise, $k = k + 1$; |
| 12. end if |
| *# Calculating the optimal strategy of the rest layers #* |
| 13. When $1 < j \le \mathcal{F}$; |
| *# Loop iteration #* |
| 14. Let $\mathbf{B}^{j+1(0)} = \mathbf{B}^{j*}$, $\mathbf{P}^{j+1(0)} = \mathbf{P}^{j*}$, and $\mathbf{r}^{j+1(0)} = \mathbf{r}^{j*}$, $\forall i \in [1\ U]$ and $\forall j \in [1\ \mathcal{F}]$; |
| 15. repeating step 3 to Step 11; |
| 16. $j = j + 1$; |
| *# Finding the optimal strategy #* |
| 17. Calculating $\mathbf{\Gamma} = \{\Gamma_1(\mathbf{B}^{1*}, \mathbf{P}^{1*}, \mathbf{r}^{1*}), ..., \Gamma_{\mathcal{F}}(\mathbf{B}^{\mathcal{F}*}, \mathbf{P}^{\mathcal{F}*}, \mathbf{r}^{\mathcal{F}*})\}$; |
| 18. $(\mathbf{s}, \mathbf{B}, \mathbf{P}, \mathbf{r}) \leftarrow \arg\min_{s^*, \mathbf{B}^*, \mathbf{P}^*, \mathbf{r}^*} \mathbf{\Gamma}$; |
| 19. If $B > 0.5 \rightarrow B = 1$; |
| 20. otherwise $B = 0$. |
| *# Finding the optimal QoE approximation #* |
| 21. If $\mathcal{R}_i > 0.5 \rightarrow \mathcal{R}_i = 1$; |
| 22. otherwise $\mathcal{R}_i = 0$. |

The Li-GD algorithm presented in TABLE I is composed of three parts.

1) **(Line2-Line12):** Calculating the optimal strategy when the model segmentation point is in the first layer. For the Li-GD algorithm, since the model segmentation strategy is discrete, we need to calculate the optimal resource allocation strategies layer by layer. For the first layer, its starting values are $\mathbf{B}^{1(0)} = \{B^{1(0)}_1, ..., B^{1(0)}_U\}$, $\mathbf{P}^{1(0)} = \{P^{1(0)}_1, ..., P^{1(0)}_U\}$ and $\mathbf{r}^{1(0)} = \{r^{1(0)}_1, ..., r^{1(0)}_U\}$, where $\mathbf{B}^{1(0)} \in [0\ 1]$, $\mathbf{r}^{1(0)} \in [r_{min}\ r_{max}]$, and $\mathbf{P}^{1(0)} \in [p_{min}\ p_{max}]$. Additionally, they are selected without any information of the final optimal values. Then, the GD algorithm is executed with step size $\lambda$ and gradient $-\mathbf{g}_k$. After k rounds of iterations, when the threshold of accuracy is reached, the optimal solutions of resource allocation strategy for the first layer are $\mathbf{B}^{1*} \leftarrow \mathbf{B}^{1(k)}$, $\mathbf{P}^{1*} \leftarrow \mathbf{P}^{1(k)}$, and $\mathbf{r}^{1*} \leftarrow \mathbf{r}^{1(k)}$.

2) **(Line13-Line16):** Calculating the optimal resource strategy for the rest layers. When the optimal resource allocation strategy of the first layer is calculated, then from the second layer, the starting values of this layer are the optimal values of the former layer whose intermediate data size is the closest with this layer. For instance, for the second layer, its starting values are $\mathbf{B}^{2(0)} = \mathbf{B}^{1*}$, $\mathbf{P}^{2(0)} = \mathbf{P}^{1*}$ and $\mathbf{r}^{2(0)} = \mathbf{r}^{1*}$. For the third layer, we compare the intermediate data size of the first and the second layers, if $|d^{\mathcal{M}}_1 - d^{\mathcal{M}}_3| > |d^{\mathcal{M}}_2 - d^{\mathcal{M}}_3|$, then $\mathbf{B}^{3(0)} = \mathbf{B}^{2*}$, $\mathbf{P}^{3(0)} = \mathbf{P}^{2*}$, and $\mathbf{r}^{3(0)} = \mathbf{r}^{2*}$, otherwise, $\mathbf{B}^{3(0)} = \mathbf{B}^{1*}$, $\mathbf{P}^{3(0)} = \mathbf{P}^{1*}$, and $\mathbf{r}^{3(0)} = \mathbf{r}^{1*}$, etc. The GD process is the same as that when calculating the optimal strategy for first layer. Therefore, in this stage, the optimal resource allocation strategies for αth ($2 \le \alpha \le \mathcal{F}$) layer are calculated: firstly, calculating $\mathcal{L}^{\mathcal{M}}_{\alpha-1} = |d^{\mathcal{M}}_\alpha - d^{\mathcal{M}}_1|$, $\mathcal{L}^{\mathcal{M}}_{\alpha-2} = |d^{\mathcal{M}}_\alpha - d^{\mathcal{M}}_2|$, ..., $\mathcal{L}^{\mathcal{M}}_{\alpha-(\alpha-1)} = |d^{\mathcal{M}}_\alpha - d^{\mathcal{M}}_{\alpha-1}|$; then, let $\alpha^* = \arg\min_{\alpha}(\mathcal{L}^{\mathcal{M}}_{\alpha-1}, \mathcal{L}^{\mathcal{M}}_{\alpha-2}, ..., \mathcal{L}^{\mathcal{M}}_{\alpha-(\alpha-1)})$; finally, setting $\mathbf{B}^{\alpha(0)} = \mathbf{B}^{\alpha*}$, $\mathbf{P}^{\alpha(0)} = \mathbf{P}^{\alpha*}$, and $\mathbf{r}^{\alpha(0)} = \mathbf{r}^{\alpha*}$.

3) **(Line17-Line22):** Finding the final optimal model segmentation strategy and resource allocation strategy. When the optimal resource allocation strategies for all the layers are calculated, which are $\mathbf{B}^* = \{\mathbf{B}^{1*}, \mathbf{B}^{2*}, ..., \mathbf{B}^{\mathcal{F}*}\}$, $\mathbf{P}^* = \{\mathbf{P}^{1*}, \mathbf{P}^{2*}, ..., \mathbf{P}^{\mathcal{F}*}\}$, and $\mathbf{r}^* = \{\mathbf{r}^{1*}, \mathbf{r}^{2*}, ..., \mathbf{r}^{\mathcal{F}*}\}$, respectively, then substituting the $\mathbf{B}^*$, $\mathbf{P}^*$, and $\mathbf{r}^*$ into (18) and obtaining $\mathcal{F}$ utility values $\mathbf{U}^* = \{\mathbf{U}^{1*}, \mathbf{U}^{2*}, \dots, \mathbf{U}^{\mathcal{F}*}\}$. Finally, finding the minimum value from $\mathbf{U}^*$, and the model split strategy and resource allocation strategy that associated with this utility value are selected as the final optimal strategy. Finally, since the value range of $\mathbf{B}$ is changed from $\{0,1\}$ to $[0\ 1]$, we give the approximate rule as: if $B > 0.5$, $B = 1$; otherwise $B = 0$.

The theoretical foundations of the Li-GD approach is that for the GD algorithm, carefully selecting the start value can decrease the complexity and speed up the convergence greatly [38]. To prove the effectiveness of the proposed Li-GD algorithm, we give the following conclusions.

### B. The properties of Li-GD algorithm

In this section, we investigate the properties of Li-GD algorithm, including convergence, complexity, and approximate error. The details are presented below.

**Corollary 2.** The Li-GD algorithm is sub-liner convergent, and the number of iterations for convergent is $K \geq \frac{2\eta[f(x_0) - f^*]}{\epsilon^2}$, where $\eta$ is the step size and $\eta \leq \frac{1}{L}$, $\epsilon$ is the threshold of accuracy.

*Proof.* Based on the conclusions in [38], the convergence of $f(x)$ has three different situations: 1) if the differentiable function $f(x)$ satisfies L-Lipschitz smooth and strong-convex simultaneously, the $f(x)$ is linear convergence; 2) if it satisfies L-Lipschitz smooth and convex, the convergence is sub-linear; 3) if the differentiable function $f(x)$ only satisfies L-Lipschitz smooth, the convergence is also sub-linear but the convergence time complexity is larger than that when it satisfies L-Lipschitz smooth and convex.

Based on (27), let $\nabla = \sum_{x=1, x \neq j}^{N} \sum_{y=1}^{U} \beta_{x,y}^k P_{x,y}^k |G_{x,y}^k|^2 + \sigma^2$, then the parts of $\Gamma_{s_i}$ relate to $B_i$ are:

$$\Gamma_{s_i}(B_i) = \sum_{i=1}^{U} \omega_T^i \left( \frac{w_{s_i}}{R_{n,i}^m} + \frac{m_i}{\Phi_{j,i}^k} \right) + \sum_{i=1}^{U} \omega_E^i \left( p_{n,i}^m \cdot \frac{w_{s_i}}{R_{n,i}^m} + P_{j,i}^k \cdot \frac{m_i}{\Phi_{j,i}^k} \right) + \omega_Q \sum_{i=1}^{U} \frac{1}{1 + e^{-a\left(\frac{T_i(s_i, B_i, P_i, r_i)}{Q_i^{\mathcal{M}}} - 1\right)}} \cdot (T_i(s_i, B_i, P_i, r_i) - Q_i^{\mathcal{M}} + 1)$$

$$= \sum_{i=1}^{U} \omega_T^i \left( \frac{w_{s_i}}{\beta_{n,i}^m \cdot \frac{B_{up}}{M} \log_2\left(1 + \frac{p_{n,i}^m |h_{n,i}^m|^2}{\sum_{v=u+1}^{U} \beta_{n,v}^m p_{n,v}^m |h_{n,v}^m|^2 + \Delta}\right)} + \frac{m_i}{\beta_{j,i}^k \frac{B_{down}}{M} \log_2\left(1 + \frac{|H_{j,i}^k|^2 P_{j,i}^k}{\sum_{q=u+1}^{U} \beta_{j,q}^k P_{j,q}^k |H_{j,q}^k|^2 + \nabla}\right)} \right) + \sum_{i=1}^{U} \omega_E^i \left( p_{n,i}^m \cdot \frac{w_{s_i}}{\beta_{n,i}^m \cdot \frac{B_{up}}{M} \log_2\left(1 + \frac{p_{n,i}^m |h_{n,i}^m|^2}{\sum_{v=u+1}^{U} \beta_{n,v}^m p_{n,v}^m |h_{n,v}^m|^2 + \Delta}\right)} + P_{j,i}^k \cdot \frac{m_i}{\beta_{j,i}^k \frac{B_{down}}{M} \log_2\left(1 + \frac{|H_{j,i}^k|^2 P_{j,i}^k}{\sum_{q=u+1}^{U} \beta_{j,q}^k P_{j,q}^k |H_{j,q}^k|^2 + \nabla}\right)} \right) + \omega_Q \sum_{i=1}^{U} \frac{1}{1 + e^{-a\left(\frac{T_i(s_i, B_i, P_i, r_i)}{Q_i^{\mathcal{M}}} - 1\right)}} \cdot (T_i(s_i, B_i, P_i, r_i) - Q_i^{\mathcal{M}} + 1) \quad (36)$$

Moreover, $T_i(s_i, B_i, P_i, r_i)$ can be expressed as:

$$T_i(s_i, B_i, P_i, r_i) = \sum_{i=1}^{U} \omega_T^i \left( \sum_{\delta=1}^{s_i} \frac{f_{l_\delta}}{c_i} + \sum_{\delta=s_i+1}^{\mathcal{F}} \frac{f_{e_\delta}}{\lambda(r_i) c_{min}} + \frac{w_{s_i}}{\beta_{n,i}^m \cdot \frac{B_{up}}{M} \log_2\left(1 + \frac{p_{n,i}^m |h_{n,i}^m|^2}{\sum_{v=u+1}^{U} \beta_{n,v}^m p_{n,v}^m |h_{n,v}^m|^2 + \Delta}\right)} + \frac{m_i}{\beta_{j,i}^k \frac{B_{down}}{M} \log_2\left(1 + \frac{|H_{j,i}^k|^2 P_{j,i}^k}{\sum_{q=u+1}^{U} \beta_{j,q}^k P_{j,q}^k |H_{j,q}^k|^2 + \nabla}\right)} \right)$$

Let:

$$A = \frac{w_{s_i}}{\beta_{n,i}^m \cdot \frac{B_{up}}{M} \log_2\left(1 + \frac{p_{n,i}^m |h_{n,i}^m|^2}{\sum_{v=u+1}^{U} \beta_{n,v}^m p_{n,v}^m |h_{n,v}^m|^2 + \Delta}\right)} \quad (37)$$

$$B = \frac{m_i}{\beta_{j,i}^k \frac{B_{down}}{M} \log_2\left(1 + \frac{|H_{j,i}^k|^2 P_{j,i}^k}{\sum_{q=u+1}^{U} \beta_{j,q}^k P_{j,q}^k |H_{j,q}^k|^2 + \nabla}\right)} \quad (38)$$

For $\Gamma_{s_i}(B_i)$, A and B are similar for uplink subchannel allocation $\beta_{n,i}^m$ and downlink subchannel allocation $\beta_{j,i}^k$, which can be simplified as:

$$f(x) = \frac{1}{x \log_2\left(1 + \frac{1}{x}\right)} \quad (39)$$

And the first-order derivative of (B.4) can be expressed as:

$$f'(x) = \frac{1}{x^2 \log_2\left(1 + \frac{1}{x}\right)} \left( \frac{1}{(1+x) \ln 2 \log_2\left(1 + \frac{1}{x}\right)} - 1 \right) \quad (40)$$

*L-Lipschitz smooth:*

Let $y = kx$ and $k > 1$, then we have $|x - y| = x|k - 1|$ and:

$$|f'(x) - f'(y)| = \left| \frac{1}{k^2 x^2 \log_2\left(1 + \frac{1}{kx}\right)} \left( \frac{1}{(1+kx) \ln 2 \log_2\left(1 + \frac{1}{kx}\right)} - 1 \right) - \frac{1}{x^2 \log_2\left(1 + \frac{1}{x}\right)} \left( \frac{1}{(1+x) \ln 2 \log_2\left(1 + \frac{1}{x}\right)} - 1 \right) \right| \quad (41)$$

Since $\frac{1}{\log_2\left(1 + \frac{1}{x}\right)}$ increases with the increasing of x and $\frac{1}{kx}$ decreases with the increasing of x, we have:

$$\frac{1}{k^2 x^2 \log_2\left(1 + \frac{1}{kx}\right)} \left( \frac{1}{(1+kx) \ln 2 \log_2\left(1 + \frac{1}{kx}\right)} - 1 \right) < \frac{1}{kx^2 \log_2\left(1 + \frac{1}{x}\right)} \left( \frac{1}{(1+x) \ln 2 \log_2\left(1 + \frac{1}{kx}\right)} - 1 \right) \quad (42)$$

Firstly, let:

$$\varphi = \frac{\frac{1}{kx^2 \log_2\left(1 + \frac{1}{x}\right)}}{\frac{1}{k^2 x^2 \log_2\left(1 + \frac{1}{kx}\right)}} = \frac{k^2 x^2 \log_2\left(1 + \frac{1}{kx}\right)}{kx^2 \log_2\left(1 + \frac{1}{x}\right)} = \frac{k \log_2\left(1 + \frac{1}{kx}\right)}{\log_2\left(1 + \frac{1}{x}\right)} \quad (43)$$

Since $0 < x \leq 1$, we have: $\log_2\left(1 + \frac{1}{x}\right) > 1$ and $\log_2\left(1 + \frac{1}{kx}\right) > 1$. Therefore:

$$\theta = \frac{k 2^{\log_2\left(1 + \frac{1}{kx}\right)}}{2^{\log_2\left(1 + \frac{1}{x}\right)}} = \frac{kx + 1}{1 + x} > 1 \quad (44)$$

Thus, (44) means $k \log_2\left(1 + \frac{1}{kx}\right) > \log_2\left(1 + \frac{1}{x}\right)$, which also means $\varphi > 1$. Moreover, it is obvious that $\frac{1}{(1+kx) \ln 2 \log_2\left(1 + \frac{1}{kx}\right)} - 1 < \frac{1}{(1+x) \ln 2 \log_2\left(1 + \frac{1}{kx}\right)} - 1$.

Therefore, (41) can be rewritten as:

$$|f'(x) - f'(y)| \leq \left| \frac{1}{kx^2 \log_2\left(1 + \frac{1}{x}\right)} \left( \frac{1}{(1+x) \ln 2 \log_2\left(1 + \frac{1}{kx}\right)} - 1 \right) - \frac{1}{x^2 \log_2\left(1 + \frac{1}{x}\right)} \left( \frac{1}{(1+x) \ln 2 \log_2\left(1 + \frac{1}{x}\right)} - 1 \right) \right|$$

$$= \frac{1}{x^2 (1+x) \ln 2 \log_2\left(1 + \frac{1}{x}\right)} \left| \frac{1}{k \log_2\left(1 + \frac{1}{kx}\right)} - \frac{1}{\log_2\left(1 + \frac{1}{x}\right)} \right|$$

$$\leq \frac{1}{x^2 (1+x) \ln 2 \log_2\left(1 + \frac{1}{x}\right)} \left| \frac{1}{\log_2\left(1 + \frac{1}{kx}\right)} - \frac{1}{\log_2\left(1 + \frac{1}{x}\right)} \right| \quad (45)$$

If the (45) is no larger than $L|x-y|$ holds, then the following constraints should be satisfied:

$$\frac{1}{x^2(1+x)\ln 2\log_2\left(1+\frac{1}{x}\right)}\left|\frac{1}{\log_2\left(1+\frac{1}{kx}\right)} - \frac{1}{\log_2\left(1+\frac{1}{x}\right)}\right|$$
$$< Lx|k-1| = L|x-y| \quad (46)$$

The (46) equals to:

$$L > \frac{1}{(k-1)x^3(1+x)\ln 2\log_2\left(1+\frac{1}{x}\right)}\left|\frac{1}{\log_2\left(1+\frac{1}{kx}\right)} - \frac{1}{\log_2\left(1+\frac{1}{x}\right)}\right| \quad (47)$$

Moreover, it is easy to be proved that $j(x) = \frac{1}{\log_2\left(1+\frac{1}{kx}\right)} - \frac{1}{\log_2\left(1+\frac{1}{x}\right)}$ is monotone increasing with x, which is presented as follows.

$$j'(x) = \frac{1}{kx^2\left(1+\frac{1}{kx}\right)\ln 2\log_2\left(1+\frac{1}{kx}\right)} - \frac{1}{x^2\left(1+\frac{1}{x}\right)\ln 2\log_2\left(1+\frac{1}{x}\right)} \quad (48)$$

Let:

$$l(x) = \frac{\frac{1}{kx^2\left(1+\frac{1}{kx}\right)\ln 2\left(\log_2\left(1+\frac{1}{kx}\right)\right)^2}}{\frac{1}{x^2\left(1+\frac{1}{x}\right)\ln 2\left(\log_2\left(1+\frac{1}{x}\right)\right)^2}} = \frac{x^2\left(1+\frac{1}{x}\right)\ln 2\left(\log_2\left(1+\frac{1}{x}\right)\right)^2}{kx^2\left(1+\frac{1}{kx}\right)\ln 2\left(\log_2\left(1+\frac{1}{kx}\right)\right)^2}$$
$$= \frac{(1+x)\left(\log_2\left(1+\frac{1}{x}\right)\right)^2}{(1+kx)\left(\log_2\left(1+\frac{1}{kx}\right)\right)^2} > \frac{\left(\log_2\left(1+\frac{1}{x}\right)\right)^2}{k\left(\log_2\left(1+\frac{1}{kx}\right)\right)^2} > \frac{\left(\log_2\left(\frac{1}{k}+\frac{1}{x}\right)\right)^2}{k\left(\log_2\left(1+\frac{1}{kx}\right)\right)^2}$$
$$= \frac{(\log_2(k+x) - \log_2(kx))^2}{k(\log_2(1+kx) - \log_2(kx))^2} = \left(\frac{\log_2(k+x)}{\log_2(1+kx)^{\sqrt{k}}}\right)^2 \quad (49)$$

Let $\eta = \sqrt{k} > 1$, then we have:
$$v(x) = k + x - (1+kx)^\eta \quad (50)$$

According to (50), we have:
$$v'(x) = 1 - \eta k(1+kx)^{\eta-1} < 0 \quad (51)$$

Therefore, for $v(x)$, when $x=0$, it has the maximum value $v(0) = k - 1 > 0$. Thus, we can conclude that $\frac{\log_2(k+x)}{\log_2(1+kx)^{\sqrt{k}}} > 1$, which means that $l(x) > 1$ and $j'(x) > 0$.

Thus, for (47), we can conclude that:

$$L > \frac{1}{2(k-1)\ln 2}\left|\frac{1}{\log_2\left(1+\frac{1}{k}\right)} - 1\right| \quad (52)$$

Since $k > 1$, $\frac{1}{2(k-1)\ln 2}\left|\frac{1}{\log_2\left(1+\frac{1}{k}\right)} - 1\right| < 1$. Therefore, according to (52), there must exist $L > 1$ can satisfy the constraint in (46). Thus, $T_i(s_i, B_i, P_i, r_i)$ is L-Lipschitz smooth. Let $f(x) = T_i(s_i, B_i, P_i, r_i)$ and $g(x) = \frac{1}{1+e^{-a(x-1)}}$, then we can prove that $g(f(x))$ and $g(f(x)) \cdot f(x)$ are all L-Lipschitz smooth.

For $g(f(x))$, we have:
$$g'(f(x)) = g'(x) \cdot f'(x) \quad (53)$$
then
$$|g'(f(x)) - g'(f(y))| = |g'(x) \cdot f'(x) - g'(y) \cdot f'(y)| \quad (54)$$
Since we already have $|f'(x) - f'(y)| < L_f \cdot |x-y|$, then () can be expressed as:
$$|g'(x) \cdot f'(x) - g'(y) \cdot f'(y)| < |g *| \cdot |f'(x) - f'(y)|$$
$$< |g *| \cdot L_f \cdot |x - y| \quad (55)$$

Let $L_g^* = |g *| \cdot L_f$, thus, there exist $L_g^* > 1$ satisfies (55). Thus, $g(f(x))$ is L-Lipschitz smooth. Let $h(x) = g(f(x))$, then $g(f(x)) \cdot f(x)$ can be expressed as $h(x) \cdot f(x)$. then, we have:
$$(h(x) \cdot f(x))' = h'(x) \cdot f(x) + h(x) \cdot f'(x) \quad (56)$$
Thus,
$$|(h(x) \cdot f(x))' - (h(y) \cdot f(y))'|$$
$$= |h'(x) \cdot f(x) + h(x) \cdot f'(x) - h'(y) \cdot f(y) - h(y) \cdot f'(y)|$$
$$= |h'(x) \cdot f(x) - h'(y) \cdot f(y) + h(x) \cdot f'(x) - h(y) \cdot f'(y)|$$
$$\leq |h'(x) \cdot f(x) - h'(y) \cdot f(y)| + |h(x) \cdot f'(x) - h(y) \cdot f'(y)|$$
$$< |f *||h'(x) - h'(y)| + |h *||f'(x) - f'(y)|$$
$$< |f *| \cdot L_f \cdot |x-y| + |h *| \cdot L_h \cdot |x-y|$$
$$= (|f *| \cdot L_f + |h *| \cdot L_h) \cdot |x-y| \quad (57)$$

Let $L *= |f *| \cdot L_f + |h *| \cdot L_h$, thus, there exist $L_g^* > 1$ satisfies (55). Thus, $g(f(x)) \cdot f(x)$ is L-Lipschitz smooth. This also means $\Gamma_{s_i}(B_i)$ is L-Lipschitz smooth. According to the same method, the $\Gamma_{s_i}$ relates to $P_i$ is the same as (36). Therefore, the $\Gamma_{s_i}(P_i)$ can be simplified to four basic functions $y(x) = \frac{1}{\log_2\left(1+\frac{1}{x}\right)}$, $g(x) = \frac{1}{\log_2(1+x)}$, $h(x) = \frac{x}{\log_2(1+x)}$, and $z(x) = \frac{x}{\log_2\left(1+\frac{1}{x}\right)}$. The same as the process that shown above, the $y(x), g(x), h(x)$, and $z(x)$ are all L-Lipschitz smooth.

*Convex:*

For the differentiable function $f(x)$, if $f''(x) > 0$, it is convex [37]. Unfortunately, since $g(x) = \frac{1}{\log_2(1+x)}$, $g''(x) < 0$. Thus, the convex of $\Gamma_{s_i}(B_i)$ cannot be guaranteed. Thus, the Corollary 2 holds. ∎

**Corollary 4.** The Li-GD algorithm can accelerate the convergence of GD algorithm while reducing complexity.

*Proof.* From Corollary 2 and Corollary 3, we can conclude that for the fixed precision $\varepsilon$ and step size $\eta$, the convergence time $K$ is corelated to the starting point $\boldsymbol{B}^{(0)}, \boldsymbol{P}^{(0)}$, and $\boldsymbol{r}^{(0)}$, and the complexity is associated with the convergence time $K$. Thus, for reducing convergence time and complexity, the starting point should be selected carefully. Moreover, for the GD algorithm, carefully selecting the start value can decrease the complexity and speed up the convergence greatly [37].

In the traditional GD algorithm, for each round of GD, the starting values are $\boldsymbol{B}^{(0)}, \boldsymbol{P}^{(0)}$, and $\boldsymbol{r}^{(0)}$. Since we do not have any information about the starting values of each round GD process, they always are set to 0. The convergence time is $K_1 = max\left\{\frac{\|\boldsymbol{B}^{(0)}-\boldsymbol{B}^*\|_2^2}{2\eta\varepsilon}, \frac{\|\boldsymbol{r}^{(0)}-\boldsymbol{r}^*\|_2^2}{2\eta\varepsilon}, \frac{\|\boldsymbol{P}^{(0)}-\boldsymbol{P}^*\|_2^2}{2\eta\varepsilon}\right\}$. However, as introduced in Section III.A, for the Li-GD algorithm, the starting values are the optimal solutions of the former layer whose intermediate data size is the closest with this layer, which are $\boldsymbol{B}^{\alpha(0)} = \boldsymbol{B}^{\alpha*}$, $\boldsymbol{P}^{\alpha(0)} = \boldsymbol{P}^{\alpha*}$, and $\boldsymbol{r}^{\alpha(0)} = \boldsymbol{r}^{\alpha*}$, where $\alpha^* = \arg\min_\alpha(\mathcal{L}_{\alpha-1}^\mathcal{M}, \mathcal{L}_{\alpha-2}^\mathcal{M}, \ldots, \mathcal{L}_{\alpha-(\alpha-1)}^\mathcal{M})$, i.e., $\boldsymbol{B}^{\alpha*}, \boldsymbol{P}^{\alpha*}$, and $\boldsymbol{r}^{\alpha*}$ are the one of the optimal result of the form $(\alpha-1)$ layers whose intermediate data size is the closest to $\alpha th$ layer. Therefore, the convergence time of Li-GD algorithm is $K_2 = max\left\{\frac{\|\boldsymbol{B}^{\alpha*}-\boldsymbol{B}^*\|_2^2}{2\eta\varepsilon}, \frac{\|\boldsymbol{r}^{\alpha*}-\boldsymbol{r}^*\|_2^2}{2\eta\varepsilon}, \frac{\|\boldsymbol{P}^*-\boldsymbol{P}^{\alpha*}\|_2^2}{2\eta\varepsilon}\right\}$. Since $|\boldsymbol{B}^{\alpha*}-\boldsymbol{B}^*|, |\boldsymbol{P}^*-\boldsymbol{P}^{\alpha*}|$, and $|\boldsymbol{r}^{\alpha*}-\boldsymbol{r}^*|$ are much smaller than $|\boldsymbol{B}^{(0)}-\boldsymbol{B}^*|, |\boldsymbol{P}^{(0)}-\boldsymbol{P}^*|$, and $|\boldsymbol{r}^{(0)}-\boldsymbol{r}^*|$, the convergence time is accelerated.

Additionally, there are $\mathcal{F}$ layers, for the traditional GD algorithm, the total convergence time is $\mathcal{F}K_1$. For Li-GD algorithm, the total convergence time is $K_1 + \sum_{j=2}^{\mathcal{F}} K_2^j$. Since $K_2^j$ is much smaller than $K_1$, the complexity is reduced. ∎

**Corollary 5.** The approximate error $\varphi$ of Li-GD algorithm is less than $\frac{\partial'\varepsilon}{\rho_{min}(1-B_i^{max})\log_2\left(1+\frac{P_{min}}{\Delta^*+\frac{\alpha P_{max}}{2}}\right)}$ with $\partial' < \left(1+e^{-a(T_i^\mathcal{M}/Q_i^\mathcal{M}-1)}\right) \cdot \frac{B_2^{max}}{B^{min}}$.

*Proof.* As shown in Corollary 1, the approximation is caused by the approximate of $\beta_{n,i}^m \in \{0,1\}$ and $\beta_{j,i}^k \in \{0,1\}$ to $\beta_{n,i}^m \in [0\ 1]$ and $\beta_{j,i}^k \in [0\ 1]$. Therefore, according to (7), we have:

$$\partial = \frac{1}{\rho_2(1-B_i^2)\log_2\left(1+\frac{P_2}{\mathbb{R}^*+\mathbb{R}'_{B_i^2}}\right)} - \frac{1}{\rho_1 B_i^1 \log_2\left(1+\frac{P_1}{\mathbb{R}^*+\mathbb{R}'_{B_i^1}}\right)} \quad (58)$$

where $B_i^2$ means that the value of $B_i$ is larger than 0.5, $\rho_2$ is the probability that $B_i > 0.5$; $B_i^1$ means that the value of $B_i$ is smaller than 0.5; $\rho_1$ is the probability that $B_i^1 < 0.5$; $\mathbb{R}^*$ is the inter-cell interference and intra-cell interference under optimal circumstance; $\mathbb{R}'_{B_i^2}$ and $\mathbb{R}'_{B_i^1}$ are the increased intra-cell interference and inter-cell interference under these two circumstances. Additionally, $0 < \rho_2 < 1$, $0.5 < B_2 < 1$, $0 < \rho_1 < 1$, and $0 < B_1 < 0.5$. Let $P_2 = \alpha P_1$, therefore, the (58) equals to:

$$\partial < \frac{1}{\rho_2(1-B_i^2)\log_2\left(1+\frac{P_2}{\mathbb{R}^*+\mathbb{R}'_{B_i^2}}\right)} - \frac{2}{\log_2\left(1+\frac{P_1}{\mathbb{R}^*+\mathbb{R}'_{B_i^1}}\right)} \quad (59)$$

Moreover, for the $\mathbb{R}'_{B_i^2}$ and $\mathbb{R}'_{B_i^1}$, we have:

$$\mathbb{R}'_{B_i} = \rho_2(1-B_i^2)P_2 - \rho_1 B_i^1 P_1 < \frac{\rho_2 P_2 - \rho_1 P_1}{2}$$
$$= \frac{P_1(\alpha\rho_2 - \rho_1)}{2} < \frac{P_{max}(\alpha\rho_2 - \rho_1)}{2} < \frac{\alpha P_{max}}{2} \quad (60)$$

Thus, (60) equals to:

$$\partial < \frac{1}{\rho_2(1-B_i^2)\log_2\left(1+\frac{P_2}{\Delta^*+\frac{\alpha P_{max}}{2}}\right)} - \frac{2}{\log_2\left(1+\frac{P_1}{\Delta^*+\frac{\alpha P_{max}}{2}}\right)}$$
$$< \frac{1}{\rho_2(1-B_i^2)\log_2\left(1+\frac{P_{min}}{\Delta^*+\frac{\alpha P_{max}}{2}}\right)} < \frac{1}{\rho_{min}(1-B_i^{max})\log_2\left(1+\frac{P_{min}}{\Delta^*+\frac{\alpha P_{max}}{2}}\right)} \quad (61)$$

Another approximation comes from the parameter $\mathcal{R}_i = \frac{1}{1+e^{-a\left(T_i^\mathcal{M}/Q_i^\mathcal{M}-1\right)}}$. For $\mathcal{R}_i$, when $\mathcal{R}_i < \frac{1}{2}$, then we approximate the value of $\mathcal{R}_i = 0$; when $\mathcal{R}_i > \frac{1}{2}$, we approximate the value of $\mathcal{R}_i = 1$. However, based on (13), when $T_i^\mathcal{M} < Q_i^\mathcal{M}$, $Q_i = 0$; when $T_i^\mathcal{M} > Q_i^\mathcal{M}$, $Q_i = T_i^\mathcal{M} - Q_i^\mathcal{M}$. According to the approximate of $\mathcal{R}_i$, the value of $Q_i$ and $Q_i^*$ will be different. Let B to represent the total sum of DCT, in $Q_i$, the B is contributed only by the situation when $T_i^\mathcal{M} > Q_i^\mathcal{M}$. However, in $Q_i^*$, it comes two aspects: $B_1 = \left(T_i^\mathcal{M} - Q_i^\mathcal{M}\right)\left(\frac{T_i^\mathcal{M}}{Q_i^\mathcal{M}} - 1\right)$ when $T_i^\mathcal{M} < Q_i^\mathcal{M}$ and $B_2 = \left(T_i^\mathcal{M} - Q_i^\mathcal{M}\right)\left(\frac{T_i^\mathcal{M}}{Q_i^\mathcal{M}} - 1\right)$ when $T_i^\mathcal{M} > Q_i^\mathcal{M}$. The values of $B_1$ and $B_2$ are all larger than 0. However, in the approximation in this paper, the $B_1$ is ignored. Moreover, even $B_2$ is calculated, since $\left(\frac{T_i^\mathcal{M}}{Q_i^\mathcal{M}} - 1\right) < 1$, the value of $\left(T_i^\mathcal{M} - Q_i^\mathcal{M}\right)$ is larger than that in $Q_i^*$. However, when we approximate $\mathcal{R}_i = 1$, the value of $B_2$ increases to $B_2^*$ and $\frac{B_2^*}{B_2} = 1 + e^{-a\left(T_i^\mathcal{M}/Q_i^\mathcal{M}-1\right)}$. Therefore, considering the value of DCT is B, then the approximate ration is $\partial' = \left(1 + e^{-a\left(T_i^\mathcal{M}/Q_i^\mathcal{M}-1\right)}\right) \cdot \frac{B_2}{B} < \left(1 + e^{-a\left(T_i^{min}/Q_i^{max}-1\right)}\right) \cdot \frac{B_2^{max}}{B^{min}}$. However, in $\partial'$, as shown in Fig., with the increasing of $a$, the $\partial'$ reduces sharply. Moreover, when the value of $a$ is large, the approximate that caused by $\mathcal{R}_i$ is small enough to be ignored.

Moreover, the accuracy of GD algorithm is $\varepsilon$, therefore, the approximate error of Li-GD algorithm is $\frac{\partial'\varepsilon}{\rho_{min}(1-B_i^{max})\log_2\left(1+\frac{P_{min}}{\Delta^*+\frac{\alpha P_{max}}{2}}\right)} \cdot \frac{B_2^{max}}{B^{min}}$ with $\partial' < \left(1 + e^{-a\left(T_i^{min}/Q_i^{max}-1\right)}\right)$. ∎

Note that the approximate error $\partial'$ that calculated in Corollary 5 could be small enough by increasing the value of $a$.

Note that the convergence time can be further reduced by optimizing the step size or by using the self-adaptive step size. Moreover, lowering the accuracy can also accelerate the convergence. Therefore, by carefully achieving a tradeoff between accuracy and convergence time also can improve the performance of Li-GD algorithm. However, these are not investigated in this paper.

## V. PERFORMANCE EVALUATION

### A. Experimental Setup

*Network and Communication set*. We deploy 5 APs and 1250 users in the network. The system bandwidth is 10 MHz, which is available for all the APs. In addition, we assume that each subchannel can be accessed by at most 3 devices. The uplink channels are all independent and identically distributed Rayleigh fading channels. Referring to [39], the number of subchannels is 250; the maximum transmission power of device is 25dBm; the circuit power consumption of each edge server is 50dBm; the path loss exponent is 5; the noise power spectral density is -174dbm/Hz; the CPU cycles for 1bit task are $10^4$ cycles/bit.

*Dataset*. We use CIFAR-10 dataset in this paper. The CIFAR-10 dataset consists of 60000 $32 \times 32$ RGB images in 10 classes (from 0 to 9), with 50000 training images and 1000 test images per class.

*DNN benchmarks*. There are many DNN models with different topologies have been proposed recently. For instance, NiN, tiny YOLOv2, VGG16, etc., are the well-known chain topology models; AlexNet, ResNet-18, etc., are the well-known DAG topology models. However, in this paper, we mainly evaluate the performance of the proposed algorithms on chain topology models, i.e., NiN (9 layers), YOLOv2 (17 layers) and VGG16 (24 layers).

*Evaluation benchmarks*. We compare the proposed algorithms against Device-Only (i.e., executing the entire DNN on the device), Edge-Only (i.e., executing the entire DNN on the edge), IAO [18], DINA [14], Neurosurgeon [40], and DNN surgeon [17]. However, the DNN surgeon and DINA can operate on both chain topology models and DAG topology models. In this paper, since we mainly focus on chain topology, we only implement these three algorithms on chain topology models (i.e., NiN, YOLOv2, and VGG16).

*Evaluation variables*. In this paper, we use the latency speedup and energy consumption reduction to evaluate the performance of the benchmarks and ECC. The latency speedup means that the times of inference latency reduction compared with the baseline algorithm. For instance, assume that the baseline algorithm is Device-Only, if the ERA algorithm is applied and the inference latency in ERA is 5 times lower than that in Device-Only, then the latency speedup of ERA is 5. The meaning of energy consumption reduction is similar to that of latency speedup. However, in different simulation, the baseline algorithm is different, which will be shown in Section V.B and Section V.C.

### B. Performance under different models

In this section, we compare the performance of ERA to that of the Device-Only, Neurosurgeon, DNN surgeon, IAO, and DINA. In this section, we use the Device-Only method as the baseline, i.e., the performance is normalized to the Device-Only method. The performance of latency speedup and energy consumption reduction of these algorithms under different models is presented in Fig.6 and Fig.7, respectively.

From Fig.6, it can be found that the latency speedup in ERA is the best. The performance in Neurosurgeon, DNN surgeon, IAO, and DINA is similar. The performance in Device-Only is the worst. This is because the entire inference task is executed on device when using the Device-Only approach, in which the computing capability is lower than the edge server. Moreover, the performance of ERA is also better than Edge-Only. This is because even the computing capability of edge server is better than the device, the large amount of raw data increases the data transmission delay. Additionally, the performance in VGG16 is better than that in NiN and YOLOv2.

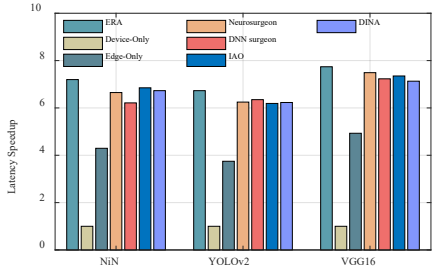

Fig.6. Latency speedup with different DNN models

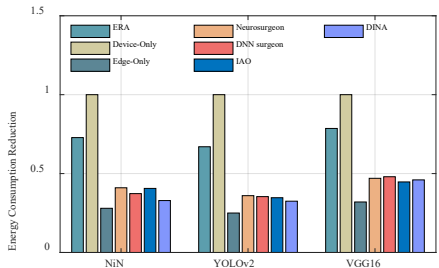

Fig.7. Energy consumption reduction with different DNN models

The performance of energy consumption reduction is shown in Fig.7. We can find that the energy consumption in Device-Only is better than the other approaches. The performance in Neurosurgeon, DNN surgeon, IAO, and DINA is similar, which are lower than that in ERA. The reason is that in ERA, Neurosurgeon, DNN surgeon, IAO, DINA and Edge-Only, on one hand, there is large amount data transmission between device and edge server, on the other hand, the power for data processing in edge server is higher than that in device. Moreover, the energy consumption reduction in VGG16 is also better than that in NiN and YOLOv2, which is similar to that in Fig.6.

*C. Performance under different QoE requirements*

In this section, we evaluate the performance of ERA under different QoE requirements, including the different QoE thresholds and different expected task finish time. Moreover, we also compare the performance of ERA with the benchmarks under different expected task finish time. The results are presented in Fig.8 to Fig.9.

*Performance under different QoE thresholds.* The performance of the latency speedup and energy consumption reduction in ERA with different models under different QoE thresholds is presented in Fig.8 and Fig.9, respectively. From Fig.8, we can find that with the decreasing of the QoE threshold, the latency speedup reduces in both these three models. For instance, when the QoE threshold is 98%, the latency speedup is about 6.9, 6.6, and 7.9 times in NiN, YOLOv2, and VGG16, respectively. When the QoE threshold reduces to 88%, these values are about 6.02, 5.8, and 7.1 times. The reason is obvious, since reducing the QoE threshold, the requirement on inference latency reduces, too, which will cause the increasing of latency. Additionally, the performance in VGG16 is better than that in NiN and YOLOv2, in which the performance is similar.

The performance of energy consumption reduction is shown in Fig.9. Different with the latency speedup in Fig.5, with the decreasing QoE threshold, the energy consumption reduces in both these three models. For instance, when the QoE threshold is 98%, the energy consumption reduction is only about 1.53, 1.46, and 1.61 times in NiN, YOLOv2, and VGG16, respectively. However, when the threshold reduces to 88%, these values become to about 2.18, 1.72, and 2.3 times. The reason is similar to that in Fig.5, because reducing the threshold can reduce latency requirements, which means less energy consumption.

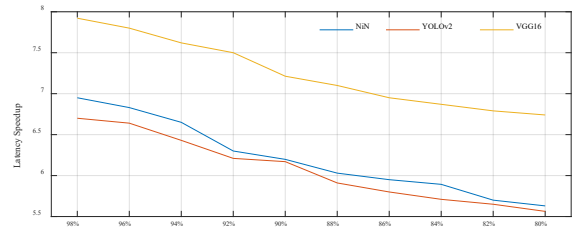

Fig.8. Latency speedup under different QoE thresholds

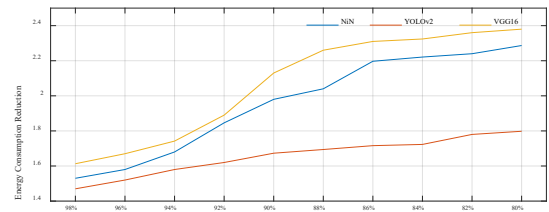

Fig.9. Energy consumption reduction under different QoE thresholds

*Performance in ERA under different expected task finish time.* The performance of ERA under different expected task finish time is presented in Fig.10 and Fig.11, respectively. From Fig.10, we can find that with the increasing of the expected finish time, the number of users whose inference delay is larger than the expected finish time decreases. In this figure, the N represents the number of users in the network and the average task finish time is 15ms. For instance, when the expected task finish time is 5ms, the number of users whose inference delay is larger than the expected finish time is about 0.67*N, 0.75*N, and 0.79*N in VGG16, NiN, and YOLOv2, respectively. However, when the expected task finish time becomes 19ms, the number of users reduces to about 0.03*N, 0.05*N, and 0.07*N. This is easy to be explained because when the expected task finish time is large, more user's inference delay will less than the expected task finish time, which means decreasing of user numbers.

The sum of inference delay in ERA under different expected task finish time with different models is presented in Fig.11. Similar to Fig.10, with the increasing of expected task finish time, the sum of inference delay that exceeds the expected task finish time decreases, too. For instance, when the expected finish time is 9ms, the sum of inference delay that exceeds the expected task finish time is about 92ms, 109ms, and 73ms in NiN, YOLOv2, and VGG16, respectively. The reason is the same as that in Fig.10.

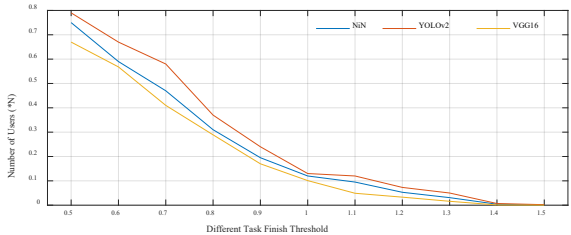

Fig.10. Number of users under different expected finish times

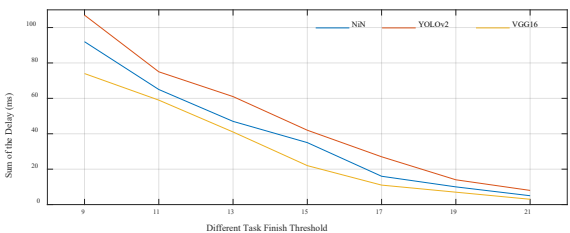

Fig.11. Sum of delay under different task finish times

*Compare ERA with other baseline approaches under different expected task finish time.* The performance of ERA, Edge-Only, Device-Only, Neurosurgeon, DNN surgeon, IAO, and DINA under different expected task finish time is presented is presented in Fig.12 and Fig.13. Because for different algorithms, the task finish time for the same task is different, therefore, in Fig.12 and Fig.13: 1) the x-axis means the percentage to the average task finish time, we use the times to the average task finish time of user to replace it; 2) the y-axis in Fig.12 means the number of users whose task finish time is larger than the expect one; the y-axis in Fig.13 means the times of the average delay over the average task finish time, i.e., the sum of the exceeds value over the expected task finish time to the average task finish time of users.

From Fig.12, we can find that with the increasing of the expected task finish time, the number of users whose inference delay is larger than the expected task finish time reduces. Additionally, for the same expected task finish time, the performance ERA is the best; the performance of DNN surgeon, Neurosurgeon, IAO, and DINA is similar, which are all better than the Edge-Only and Device-Only approaches. The performance of Edge-Only is a litter better than the Device-Only approach. Moreover, with the increasing of the excepted task finish time, the number of users whose task finish time is larger than the excepted task finish time reduces sharply. For instance, when the excepted task finish time reduces from 0.6 times to 1.2 times, the number of users in ERA reduces from 0.58*N to 0.02*N. This is easy to be understood because increasing the expected task finish time, more and more users can satisfy the QoE requirements, which means the number of users whose task finish time is larger than the threshold will be reduced.

The average inference delay that exceeds the excepted task finish time is presented in Fig.13. in Fig.13, we can conclude that with the increasing of task finish threshold, the average inference delay reduces sharply, which is similar to that in Fig.12. For instance, when the task finish threshold reduces from 0.6 times to 1.2 times, the average inference delay reduces from 5.1 times to 0.2 times in ERA. The reason is the same as that shown in Fig.13.

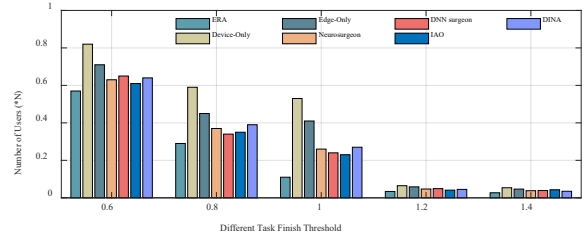

Fig.12. Number of users under different task finish threshold

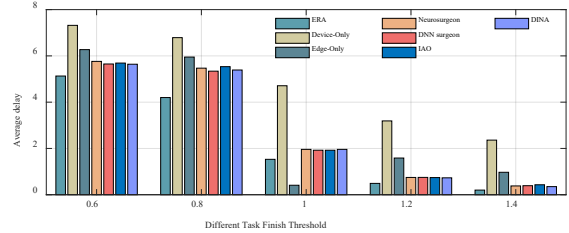

Fig.13. Average inference delay under different task finish threshold

### D. Performance under different network conditions

In this section, we compare the performance of ERA with Device-Only, Edge-Only, Neurosurgeon, DNN surgeon, IAO, and DINA under different network conditions, including different densities of mobile users (average number of users in each edge server), different number of subchannels, and different workloads. The results are presented in Fig.14 to Fig.19, respectively. In this section, we use the Device-Only method as the baseline, i.e., the performance is normalized to the Device-Only method.

The latency of ERA, Device-Only, Edge-Only, Neurosurgeon, DNN surgeon, IAO, and DINA under different user densities is presented in Fig.14. We can find that with the increasing of the user density, the latency increases in these algorithms except for Device-Only approach. Since in Device-Only approach, the whole inference model is in the mobile device, it is not affected by the variation of user density. The latency performance of ERA is the best, which is a litter better than the other algorithms with the increasing of user density. For instance, when $U = 100$, the latency speedup of ERA and Neurosurgeon is 8.9 and 8.5, respectively; when $U = 200$, this becomes 7.2 and 6.6, respectively. The reason is that the tradeoff between QoE and resource consumption are considered in ERA.

The performance of latency speedup under different number of subchannels is presented in Fig.15. We can find that with the increasing of the number of subchannels, the latency speedup of ERA increases first then reducing after $M = 100$. This is because when the number of subchannels increases, the bandwidth of each subchannel reduces. Even the number of users in each subchannel reduces, if the bandwidth of each subchannel is small enough, the data transmission rate will be reduced seriously. The performance of Edge-Only,

Neurosurgeon, DNN surgeon, IAO, and DINA are not affected because they do not use the NOMA channel. Moreover, the performance of ERA is better than the other approaches. The reason is the same as that in Fig.14.

The performance of latency speedup under different workloads is presented in Fig.16, where $k$ means the average number of works in each mobile user. In Fig.16, with the increasing of workloads, the latency in mobile devices increases, too. Therefore, in this Section, we use the latency of mobile devices when $K = 500$ as the baseline. Therefore, with the increasing of workloads, the latency increases in both these seven algorithms. The latency speedup of ERA is better than the other algorithms. The reasons are similar with that in Fig.7 and Fig.15.

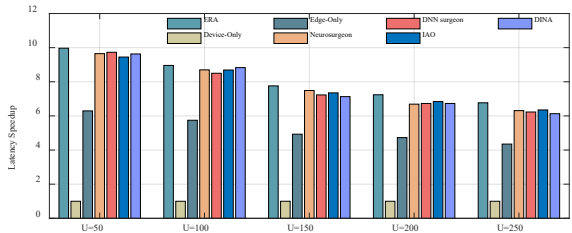

Fig.14. Latency speedup under different user densities

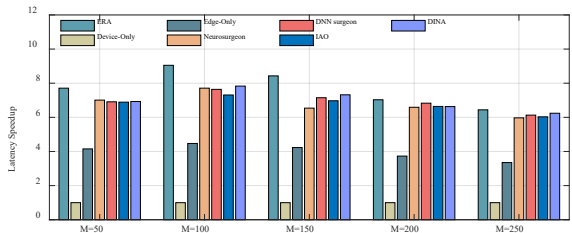

Fig.15. Latency speedup under different number of subchannels

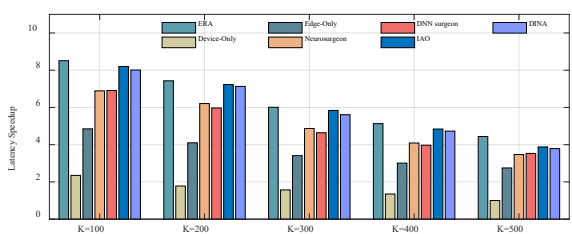

Fig.16. Latency speedup under different workload

The energy consumption of ERA, Device-Only, Edge-Only, Neurosurgeon, DNN surgeon, IAO, and DINA under different user densities is presented in Fig.17. We can find that with the increasing of the user density, the energy consumption increases in these algorithms except for Device-Only approach. Since in Device-Only approach, the whole inference model is in the mobile device, its energy consumption is not affected by the variation of user density. The energy consumption reduction of ERA is the best. The advantage of ERA becomes smaller with the increasing of the user density. This is because with the increasing of the user density, more users will share the same subchannel, for maintain high data transmission rate, the transmission power should be improved in NOMA.

The energy consumption reduction under different number of subchannels is presented in Fig.18. We can find that with the increasing of the number of subchannels, the energy consumption of ERA reduces. This is because when the number of subchannels increases, the number of mobile users in each subchannel reduces. Then the intra-cell interference reduces. Therefore, the user and edge server can lower their transmission power. However, the increasing of the number of subchannels will cause the increasing of latency, which will contribute to the increasing of energy consumption. However, since the ERA takes both the resource consumption and QoE into account, the energy consumption in ERA reduces. The performance of Edge-Only, Neurosurgeon, DNN surgeon, IAO, and DINA are not affected because they do not use the NOMA channel. Moreover, the performance of Device-Only is the best, the reasons are the same with that in Fig.7.

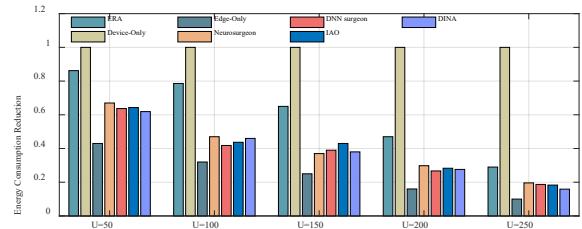

Fig.17. Energy consumption reduction under different user densities

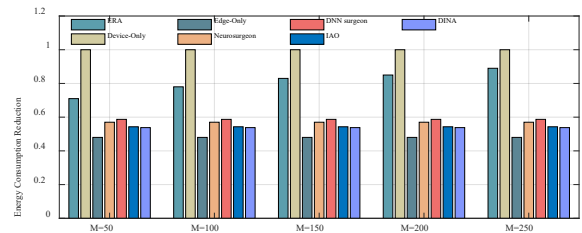

Fig.18. Energy consumption reduction under different number of subchannels

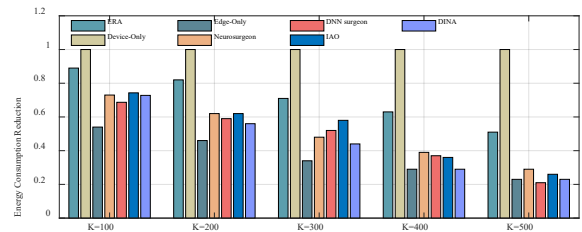

Fig.19. Energy consumption reduction under different workload

The performance of energy consumption under different workloads is presented in Fig.19. In Fig.19, with the increasing of workloads, the energy consumption in mobile devices increases, too. Therefore, in this figure, we use the latency of mobile devices when $K = 500$ as the baseline. Therefore, with the increasing of workloads, the energy consumption increases in both these seven algorithms. The energy consumption reduction of ERA is better than the other algorithms. Moreover, the advantage of ERA becomes obviously with the increasing of workload. The reasons are similar with that in Fig.16.

## VI. CONCLUSION

In this paper, considering that the previous edge split inference mainly concentrates on improving and optimizing the system's QoS, ignore the effect of QoE which is another critical item for the users except for QoS. For accelerating split inference in EI and achieving the tradeoff between inference delay, QoE, and resource consumption, we propose A QoE-Aware Split Inference Accelerating Algorithm for NOMA-based Edge Intelligence, abbreviated as ERA. Specifically, the ERA takes the resource consumption, QoE, and inference latency into account to find the optimal model split strategy and resource allocation strategy. Since the minimum inference delay and resource consumption, and maximum QoE cannot be satisfied simultaneously, the gradient descent (GD) based algorithm is adopted to find the optimal tradeoff between them. Moreover, the loop iteration GD approach (Li-GD) is developed to reduce the complexity of the GD algorithm caused by parameter discretization. Additionally, the properties of the proposed algorithms are investigated, including convergence, complexity, and approximation error. The experimental results demonstrate that the performance of ERA is much better than that of the previous studies.

ACKNOWLEDGMENT

This work was supported in part by the grant from NSFC Grant no. 62101159, NSF of Shandong Grant no. ZR2021MF055, and also the Research Grants Council of Hong Kong under the Areas of Excellence scheme grant AoE/E-601/22-R.